\RequirePackage{fix-cm}
\documentclass{svjour3}
\smartqed
\usepackage{textcomp}
\usepackage[round]{natbib}
\usepackage{algorithmic}
\usepackage{algorithm}
\usepackage{amsmath}
\usepackage{amssymb}
\usepackage{subcaption}
\usepackage{lipsum}
\usepackage{mathtools}
\usepackage{graphicx}
\usepackage[table,xcdraw]{xcolor}
\usepackage{amsfonts}
\usepackage{multirow, booktabs}
\usepackage{siunitx}
\usepackage{longtable}
\usepackage{makecell, boldline}
\usepackage{epstopdf}

\begin{document}
\newcommand*{\everymodeprime}{\ensuremath{\prime}}

\title{Cognitive-aware Short-text Understanding for Inferring Professions}

\author{Sayna Esmailzadeh         \and
	Saeid Hosseini \and
	Mohammad Reza Kangavari \and
	Wen Hua
}

\institute{Sayna Esmailzadeh \at
	Computational cognitive model research lab., School of computer engineering, Iran University of Science and Technology, Iran. \\
	\email{sayna.esmailzadeh@gmail.com}
	\and
	Saeid Hosseini \at
	Faculty of Computing and Information Technology,
	Sohar University, Sohar, Oman. \\
	\email{sahosseini@su.edu.om}
	\and
	Mohammad Reza Kangavari \at
	Computational cognitive model research lab., School of computer engineering, Iran University of Science and Technology, Iran. \\
	\email{kanagvari@iust.ac.ir}
	\and
	Wen Hua \at
	School of Information Technology and Electrical Engineering, University of Queensland, Brisbane, Australia. \\
	\email{w.hua@uq.edu.au}
	\and 
	Sayna Esmailzadeh and Saeid Hosseini are co-first authors
}
\vspace{-4mm}
\date{Received: date / Accepted: date}
\maketitle
\vspace{-4mm}
\begin{abstract}
Leveraging short-text contents to estimate the occupation of microblog authors has significant gains in many applications. Yet challenges abound. Firstly brief textual contents come with excessive lexical noise that makes the inference problem challenging. Secondly, cognitive-semantics are not evident, and important linguistic features are latent in short-text contents. Thirdly, it is hard to measure the correlation between the cognitive short-text semantics and the features pertaining various occupations. We argue that the multi-aspect cognitive features are needed to correctly associate short-text contents to a particular job and discover suitable people for the careers. To this end, we devise a novel framework that on the one hand, can infer short-text contents and exploit cognitive features, and on the other hand, fuses various adopted novel algorithms, such as curve fitting, support vector, and boosting modules to better predict the occupation of the authors. The final estimation module manufactures the $R^w$-tree via coherence weight to tune the best outcome in the inferring process. We conduct comprehensive experiments on real-life Twitter data. The experimental results show that compared to other rivals, our cognitive multi-aspect model can achieve a higher performance in the career estimation procedure, where it is inevitable to neglect the contextual semantics of users.
\vspace{-2mm}
\end{abstract}
\vspace{-6mm}
\section{Introduction}
\label{sec:intro}
\vspace{-4mm}
Cognitive-semantic approaches \citep{singh2018cognitive} combine knowledge from linguistic analytics with intellectual perceptions. Given lexical contents $t_j$ generated by the author $a_m$, we aim to estimate an occupation $o_n$ through analyzing the cognitive features. This can directly reveal the best person who is mentally fit to handle a particular task. Nowadays, social networks record brief textual contents of the authors that are generated in a high throughput rate. Inferring professions from short-text contents in a cognitive-semantic fashion finds important applications in numerous domains: (i) In crowd-sourcing \citep{sheehan2018crowdsourcing}, the career module can choose the right crowd to carry out the task \citep{navajas2018aggregated, brown2019wisdom, daniel2018quality}. (ii) In job recommendation systems, the cognitive module can reflect personality characteristics \citep{hasan2018excessive}. (iii) In workload management, the model can evaluate the suitability of the workforce, which can further improve the efficiency of the team \citep{dias2018development}.
Inherently, compared to the comprehensive formal documents, individuals tend more frequently to compose informal short-text contents, an easier approach. While the short-text contents can better reveal the linguistic features, we can leverage such informative content to estimate the suitability of professions in real-world scenarios. We address the NP-hard \citep{benabderrahmane2017smart4job} problem of associating the latent cognitive-semantics to the professions in two steps. Firstly, we exploit the linguistic features (e.g. \textit{anger}) and enhance them with cognitive features (e.g. \textit{agreeableness}) to generate the cognitive-semantic vector for each individual. Secondly, we devise a heterogeneous combination of machine learning algorithms to fit the author vectors to the relevant cluster of occupations, where the novel modules are adjusted using the coherence scores. Nevertheless, understanding the professions by cognitive-semantic cues poses certain challenges that we elucidate as follows:\\
\textit{\textbf{Challenge 1 (Excessive Lexical Noise)}} \\
\label{intro:challenge_lexical_noise}
\indent Short-text contents are limited in size and include excessive lexical noise such as various abbreviations, and spelling errors, which makes them scientifically tedious to handle \citep{hosseini2017location, hua2015short}. We improve the segmentation algorithm based on the Term Graph (TG) \citep{hua2015short}, where we consider coherence weight for each term $e_t$. Also, we adapt Symmetric Conditional Probability (SCP) \citep{hosseini2014location} to evaluate the phrases of up to 5-grams.\\
\textit{\textbf{Challenge 2 (Hidden Linguistic Features)}}\\
\label{intro:hidden_linguistic_features}
\indent Since the words may not carry pertinent instinct meanings, the genuine linguistic features turn latent in short-text contents \citep{wu2020effects, farnadi2013recognising}. To this end, cognitive processes are required to reveal the hidden concepts behind the semantics.
\begin{table}[h]
	\centering
	\def\arraystretch{1.5}
	\caption{Cognitive-semantics}
	\label{tbl:intro_linguistic_challenge}
	\vspace{-3mm}
	\begin{tabular}{|l|l|l|l|}
		\hline
		\textbf{Word} & \textbf{anger} & \textbf{fear} & \textbf{joy}                                                                                                                                                                        
		\\ \hline
		\begin{tabular}[c]{@{}l@{}}{devil, lonely, madness, psychosis, subjugation}\end{tabular} &
		\begin{tabular}[c]{@{}l@{}}1\end{tabular}   & \begin{tabular}[c]{@{}l@{}}1\end{tabular} &
		\begin{tabular}[c]{@{}l@{}}0\end{tabular}
		\\ \hline				
		\begin{tabular}[c]{@{}l@{}}{powerful, cash, excite, honest, intense}\end{tabular} &
		\begin{tabular}[c]{@{}l@{}}1\end{tabular}   & \begin{tabular}[c]{@{}l@{}}1\end{tabular} &
		\begin{tabular}[c]{@{}l@{}}1\end{tabular}
		\\ \hline
		\begin{tabular}[c]{@{}l@{}}{prune, abstract, account, exercise, incorrect}\end{tabular} &
		\begin{tabular}[c]{@{}l@{}}0\end{tabular}   & \begin{tabular}[c]{@{}l@{}}0\end{tabular} &
		\begin{tabular}[c]{@{}l@{}}0\end{tabular}
		\\ \hline
		\begin{tabular}[c]{@{}l@{}}{purify, absolution, amusing, bride, buddy}\end{tabular} &
		\begin{tabular}[c]{@{}l@{}}0\end{tabular}   & \begin{tabular}[c]{@{}l@{}}0\end{tabular} &
		\begin{tabular}[c]{@{}l@{}}1\end{tabular}
		\\ \hline
		\begin{tabular}[c]{@{}l@{}}{swim, confidence, elevation, highest, immerse}\end{tabular} &
		\begin{tabular}[c]{@{}l@{}}0\end{tabular}   & \begin{tabular}[c]{@{}l@{}}1\end{tabular} &
		\begin{tabular}[c]{@{}l@{}}1\end{tabular}
		\\ \hline
	\end{tabular}
	\vspace{-7mm}
\end{table}
Based on the cognitive observations in Table \ref{tbl:intro_linguistic_challenge}, the recognition of the meanings can only be comprehensible if the pragmatic human understanding is noted. For instance, while the word \textit{lonely} can reveal the cognitive features of anger and fear, the term \textit{buddy} can highlight the joy in a cognitive perception. Hence, to infer the professions in short-text contents, we should consider the cognitive understanding of the proposed framework.\\
\noindent\textit{\textbf{Challenge 3 (one-to-many correlation between cognitive and job entities)}}\\
\label{intro:challenge_inferring_module}
\indent	Furthermore, one aspect of the challenge is that the majority of cognitive features are partially oriented toward multiple professions. Hence, we should devise a model that can infer the pertinence between cognitive features and particular jobs using cognitive insights. To this end, We firstly extract linguistic features through adopting cognitive textual analytics \citep{iatan2017predicting, al2020comparative} \textit{(e.g. LIWC)} and subsequently, prospect the cognitive features from linguistic features through utilizing the correlation coefficients.\\
Recent works in inferring professions \citep{hu2016language, tornroos2019relationship, thakur2020correlational} propose the textual relevance between the jobs and the cognitive features. However, they ignore the hidden linguistic features in the microblog noisy context. Furthermore, even though some of the current approaches \citep{brynjolfsson2018can, zhang2016ensemble} utilize various machine learning techniques to estimate occupations, they fail to tackle the problem from various aspects, relying only on Na\"ive ensemble algorithm \citep{zhang2016ensemble}).\\
\noindent\textit{\textbf{Challenge 4 (non-linear and multiplex cognitive features)}}\\
\indent Inherently, cognitive features involve multiple non-linear characteristics. Therefore the mathematical modeling of heterogeneous approaches like support vector machine, support vector clustering, and Gaussian kernel approaches to specify the occupation boundaries \citep{shafiabady2016using} turns non-trivial.
Where Lu et al. \citep{lu2019nonparametric} proposes the Gradient Boosting modules to handle non-parametric inputs, but they do not optimize the Mean Squared Error (MSE). Other similar work \citep{cui2018effect} simultaneously track the feature-specific variations on non-linear input, ignoring the group alignment of the features toward exclusive careers. Hence, we propose a multiplex unified framework that employs three machine learning techniques: SVM-Behavior, Gradboosting, and Isotonic Curve-Fitting module. Such a heterogeneous framework can handle the non-linear cognitive-semantic input from three perspectives. The \textit{SVM-Behavior} module constructs the margins for the occupation clusters based on the input cognitive features. The Support Vector Cluster (SVC) \citep{shafiabady2016using} maps the textual vectors to improve the dimension of the job-specific spaces which further utilizes the Gaussian behavior kernel. The \textit{Gradboost behavior} module infers the professions using the \textit{Boosting Behavior Tree}. The tree model can adapt the dimension for the occupation cognitive features by dividing and mapping the orients into a binary tree. The \textit{Curve Fitting} module\citep{liu2018curve} aims to train a mathematical model to effectively estimate the jobs using various cognitive features. Fitted curves can successively visualize the correlation between the cognitive features and each of the inferred professions.\\
The proposed framework in this paper explores the cognitive cues by extracting the semantic-linguistic features. The multi-component approach employs three aspects of clustering, boosting, and curve fitting to infer the correlation between the cognitive features and professions, which will be collectively adjusted via the coherence weight. We further devise a novel $R^w$-tree structure that applies quest, update, and insert operations to distinguish occupational boundaries based on the trained weights. In short, we leverage dataset-wide shreds of evidence from noisy short-text contents to exploit the cluster of occupation boundaries. To the best of our knowledge, we propose the first study on inferring professions from microblogs via a cognitive-semantic approach. Our contributions are fourfold:
\vspace{-2mm}
\begin{itemize}
	\item[$\bullet$] We propose an effective segmentation model which firstly utilizes the term-specific cognitive features and computes the correlation scores between each pair of the words, and secondly employs an altered stickiness algorithm to extract cognitive-aware multi-words from short-text contents.
	\item[$\bullet$] We devise a heterogeneous cognitive-semantic model named as LESSN to extract hidden linguistic features and also enhances cognitive-textual analytics to reveal career-related features.
	\item[$\bullet$] We propose a multi-component unified framework that can effectively bound the professions to the microblog authors in a cognitive-semantic manner. The experimental results verify the advantage of the recommended solution over other contemporary competitors.
	\item[$\bullet$] We develop a novel $R^w$-tree structure that is consistent with the agglomerative characteristic of the professions where the block indexes in the tree can accommodate the relevant cognitive features for each occupation in the hierarchy.
\end{itemize}
\vspace{-9mm}
\subsection{Paper Organization}
\vspace{-4mm}
We organize the rest of this paper as follows: in Sec. \ref{sec:relWork}, we study the literature; in Sec. \ref{sec:problemStatement}, we clarify the problem of estimating occupations based on predictive analysis of cognitive cues and elucidates the framework overview; We describe the proposed approaches and experiments in Sec. \ref{sec:methodology} and Sec. \ref{sec:Experiment} respectively. Finally, we conclude this paper and discuss future work in Sec. \ref{sec:conclusion}.
\vspace{-5mm}
\section{Related Work}
\vspace{-4mm}
\label{sec:relWork}
As briefed in Table \ref{table:literature}, the related work comprises contextual semantics, inferring professions, and cognitive perspective.
\vspace{-6mm}
\begin{table}[h]
	\centering
	\caption{Literature}
	\label{table:literature}
	\vspace{-4mm}
\begin{tabular}{|l|l|l|}
	\hline
	Category                                & Approach                                  & Reference                                            \\ \hline
	\multirow{18}{*}{Contextual Semantics}  & \multirow{4}{*}{Semantic-Oriented}        & \citep{saif2016contextual}           \\ \cline{3-3} 
	&                                           & \citep{hua2016understand}            \\ \cline{3-3} 
	&                                           & \citep{najafipour2020soulmate}       \\ \cline{3-3} 
	&                                           & \citep{hua2015short}                 \\ \cline{2-3} 
	& \multirow{5}{*}{Short-text Inference}     & \citep{najafipour2020soulmate}       \\ \cline{3-3} 
	&                                           & \citep{hua2016quality}               \\ \cline{3-3} 
	&                                           & \citep{shirakawa2015wikipedia}       \\ \cline{3-3} 
	&                                           & \citep{li2017learning}               \\ \cline{3-3} 
	&                                           & \citep{hua2015short}                 \\ \cline{2-3} 
	& \multirow{3}{*}{Noisy Contents}           & \citep{hosseini2017location}         \\ \cline{3-3} 
	&                                           & \citep{hua2016quality}               \\ \cline{3-3} 
	&                                           & \citep{shirakawa2015wikipedia}       \\ \cline{2-3} 
	& \multirow{3}{*}{Bag-of-Words}             & \citep{shirakawa2015wikipedia}       \\ \cline{3-3} 
	&                                           & \citep{najafipour2020soulmate}       \\ \cline{3-3} 
	&                                           & \citep{mikolov2013efficient}         \\ \cline{2-3} 
	& \multirow{3}{*}{Neural Networks}          & \citep{shen2014learning}             \\ \cline{3-3} 
	&                                           & \citep{graves2012supervised}         \\ \cline{3-3} 
	&                                           & \citep{najafipour2020soulmate}       \\ \hline
	\multirow{9}{*}{Inferring Professions}  & \multirow{2}{*}{Crowdsourcing}            & \citep{sheehan2018crowdsourcing}     \\ \cline{3-3} 
	&                                           & \citep{luz2015survey}                \\ \cline{2-3} 
	& \multirow{7}{*}{Job recommendation}       & \citep{huang2016exploiting}          \\ \cline{3-3} 
	&                                           & \citep{benabderrahmane2017smart4job} \\ \cline{3-3} 
	&                                           & \citep{musto2016learning}            \\ \cline{3-3} 
	&                                           & \citep{zhang2016ensemble}            \\ \cline{3-3} 
	&                                           & \citep{schnitzer2019preselection}    \\ \cline{3-3} 
	&                                           & \citep{beel2016paper}                \\ \cline{3-3} 
	&                                           & \citep{alonso2019robust}             \\ \hline
	\multirow{8}{*}{Cognitive Perspective} & \multirow{3}{*}{cognitive features}     & \citep{lent2019social}               \\ \cline{3-3} 
	&                                           & \citep{lent2016applying}             \\ \cline{3-3} 
	&                                           & \citep{kim2019understanding}         \\ \cline{2-3} 
	& \multirow{5}{*}{Cognitive Analysis Tools} & \citep{tadesse2018personality}       \\ \cline{3-3} 
	&                                           & \citep{ong2017personality}           \\ \cline{3-3} 
	&                                           & \citep{iatan2017predicting}          \\ \cline{3-3} 
	&                                           & \citep{farnadi2013recognising}       \\ \cline{3-3} 
	&                                           & \citep{al2020comparative}            \\ \hline
\end{tabular}
\vspace{-6mm}
\end{table}
\subsection{Contextual semantics}
\vspace{-4mm}
Contextual semantics \citep{saif2016contextual} can be derived using word co-occurrences and sentiment weighting. The principle behind the concept of contextual semantics comes from the dictum-“You shall know words by the job it keeps!” (Firth, 1930- 1955). Saif et al. \citep{saif2016contextual} design a novel lexicon-based approach using a contextual representation of words from Twitter, called SentiCircles, which can capture the hidden semantics of the words from their co-occurrence patterns and subsequently updates pertinent sentiments. However, they utilize supervised learning approaches that require training data for sentiment classifier learning.
Wen Hua et al. \citep{hua2016understand, hua2015short} harvest lexical-semantic relationships between the terms by applying a probabilistic network on the web corpus. To conduct type detection, they introduce Chain and Pairwise models which combine the effects of lexical. For the disambiguation task, the weighted-vote algorithm determines the most appropriate concept for an instance. 
Moreover, \citep{najafipour2020soulmate} leverages the semantic vector space to enrich short-text contents and subsequently infer the contextual (contents+concepts) links between authors. However, due to the excessive noise in contents \citep{hosseini2017location, hua2016quality, shirakawa2015wikipedia}, the procedure for short-text understanding seems quite challenging. 
The current approach \citep{mikolov2013efficient} employs continuous Bag-of-Words (CBOW) to learn the underlying word representations through neural networks. CBOW retrieves the representation of surrounding words with the middle word. CBOW model is determined to be helpful to the understanding of the textual contents. The results for CBOW can be improved by increasing the size of the training dataset and adjusting a better choice for dimensionality.
Furthermore, Deep Neural Network (DNN) models such as Convolutional Neural Network (CNN) \citep{shen2014learning} and Recurrent Neural Network (RNN) \citep{graves2012supervised} are used to learn the low-dimensional semantic vectors for the query authors.
\citep{shen2014learning} presents a Convolutional Deep Structured Semantic Model (C-DSSM) to exploit semantical relevance through similar vectors in the contextual feature space.
Graves et al. \citep{graves2012supervised} convert the syllable orders into word sequences and apply RNN to learn the contextual representations of the input sequences.
We employ the cognitive aspect of the semantics to infer career information from short-texts. To this end, we firstly extract linguistic features. We then train a multi-component heterogeneous model to computes the similarity between cognitive cues and the ground truth, the occupation tags in the microblog contents.		
\vspace{-10mm}
\subsection{Inferring professions}
\vspace{-4mm}
Crowdsourcing \citep{sheehan2018crowdsourcing, luz2015survey} is a portmanteau of crowd and outsourcing that distributes the tasks between members on the internet. Nowadays, crowdsourcing platforms (e.g. Amazon Mechanical Turk, and Freelancer) addresses microtasks (e.g image tagging) that are difficult for computers or expensive to be handled by experts. \citep{luz2015survey} detects trustworthy workers through considering the social positions and the context of the tasks \citep{musto2016learning}.
Job recommendation models \citep{schnitzer2019preselection, zhang2016ensemble} match the user preferences to the jobs.
Zhang et al. \citep{zhang2016ensemble} propose suitable jobs for candidates based on their locations, career levels, and roles. They utilize an ensemble approach through combining of Collaborative Filtering (CF) and Content-Based Filtering (CBF) modules. While the CF model \citep{alonso2019robust} works based on the ratings of similar users, the CBF method \citep{beel2016paper} exploits item-to-item correlations using a continuous bag of the words (CBOW) \citep{najafipour2020soulmate} or continuous skip-gram, which produce a distributed representation of words.
Schnitzer et al. \citep{schnitzer2019preselection} propose a personalized job recommendation system that applies word embedding to categorize similar job descriptions. This can significantly reduce the target space for the implementation of the personalized classifier and subsequently improve the real-time recommendation.
Smart4Job \citep{benabderrahmane2017smart4job} leverages domain knowledge analytics besides a temporal predictor to provide adequate job boards for the dissemination of a new suggestion. The domain knowledge analysis focuses on the experts in the field and the semantic clarification of job boards relies on textual analytics on a controlled set of vocabularies. A job board is a website that deals specifically with professions or careers (e.g. Linkedin, Indeed ).
Similarly, we devise a diligent career estimation system that leverages various machine learning modules.
Our proposed framework differs from current works that ignore the relevance between occupations and cognitive cues.
\vspace{-4mm}
\subsection{Cognitive Perspective}
\vspace{-4mm}
Exploiting the cognitive features from textual contents can be used to identify the thoughts, behaviors, and relationships between individuals and subsequently can associate each person to a suitable job \citep{lent2019social}. The cognitive-behavioral model \citep{kim2019understanding} considers both behavioral and cognitive metrics to promote the job recommendation process. Lents et al. \citep{lent2016applying} extract an array of adaptive behaviors that people employ to adjust and thrive within educational and work environments during their occupational lifespan. In other words, \citep{lent2016applying} utilizes the career self-management (CSM) model on the cognitive features to analyze social cognitive variables including self-efficacy and outcome expectations.
cognitive features \citep{thakur2020correlational} comprise five main dimensions (Openness, Conscientiousness, Extroversion, Agreeableness, Neuroticism) \citep{tornroos2019relationship}. Tadesse et al. \citep{tadesse2018personality}, utilize Linguistic Inquiry and Word Count (LIWC) dictionary to extract 85 cognitive-linguistic features. 
LIWC-based models \citep{iatan2017predicting, ong2017personality} apply the word frequency approach to extract pronouns and identify the psychological and individualized categories.
Also, they employ Structured Programming for Linguistic Cue Extraction (SPLICE) to extract 74 linguistic features. SPLICE a web-based research tool for calculating linguistic cue values based on dictionaries, part-of-speech tagging, indices. 
Moreover, they utilize Social Network Analysis (SNA) to investigate the social structure and node interactions. SNA \citep{farnadi2013recognising} infers the personality specifications of the users to study five factors in social networks that are, network-size \citep{lin2012sharing}, betweenness \citep{seibert2001social}, density \citep{lin2012sharing}, brokerage \citep{seibert2001social}, and transitivity \citep{aghagolzadeh2012transitivity}.
Aligned with current works, we collectively consider the emotional and sentiment cues in short-text contents to extract cognitive-linguistic features.
Furthermore, we analyze Emoji characters that are more appealing and convenient for authors to convey their emotions.
\vspace{-6mm}
\section{Problem Statement}
\label{sec:problemStatement}
\vspace{-4mm}
We elucidate preliminary concepts, notations, and the framework overview.
\vspace{-8mm}
\subsection{Preliminary concepts}
\label{subsec:preliminary_concepts}
\vspace{-4mm}
\subsubsection{Definition 1 (short-text content)} 
\label{def:short-text_content}
\vspace{-4mm}
Each short-text content in the corpus $t_j \in \mathbb{T}$ has an identity ($t_j$), associated author $a_m \in \mathbb{A}$, and pertinent occupation ($t_j.o_n$). Accordingly, $\mathbb{T} (\mathbb{T} = \{T_1, T_2, ..., T_m\})$ includes all short-text contents, where $T_m$ delineates the set of short-text contents that are owned by the author $a_m$, where each $t_j$ contains multiple terms $e_t \in t_j$.
\vspace{-4mm}
\subsubsection{Definition 2 (linguistic feature)}
\label{def:linguistic_feature}
\vspace{-4mm}
The cognitive-semantics can be interpreted by the conceptual structure. There each conceptual structure can represent a linguistic feature $l_k \in \mathbb{L}$. Each short-text content $t_j$ contains several linguistic features $t_j.L_k = \{k=1, ..., i|(l_k, w_k)\}$, where $l_k$ represents the linguistic feature that gain several terms $e_t \in t_j$ and $w_k$ denotes the the weight of $l_k$. Each term $e_t$ belongs to one or more linguistic features. The more the number of terms belonging to a linguistic feature, the higher the weight of the linguistic feature will be. Take \textit{"I'm happy you came to visit my gallery."} as a sample short-text. we can analyze textual content to gain linguistic features for this sentence $\{(positive-sentiment, 0.37), (joy, 0.70), (personal-pronoun, 0.69),  (negative-emotion, 0.00),...\}$.
\vspace{-4mm}
\subsubsection{Definition 3 (cognitive feature)}
\label{def:cognitive feature}	
\vspace{-4mm}		
The associated linguistic features can be interpreted by multiple cognitive features, where features represent an author cognitive features $p_q \in \mathbb{P}$ that the $\mathbb{P}$ refers to the five-factor traits that can be mapped to $t_j.P_q = \{q=1, ..., i|(p_q, c_q)\}$. Each $p_q$ represents the cognitive features and $c_q \in [-1, 1]$ the correlation range of $p_q$. The $c_q$ of -1 indicates a perfect negative linear relationship between linguistic feature $l_k$ and cognitive feature $p_q$ and the value of 1 specifies an absolute positive linear relationship between variables.
Take "I'm happy you came to visit my gallery." short-text content as an example. we can analyze linguistic features to exploit the cognitive set as:
$\{(openness, 0.83), (extroversion, 0.86), $\\$(neuroticism, -0.81), ...\}$.
\vspace{-4mm}
\subsubsection{Definition 4 (occupation corpus)}
\label{def:occupation}
\vspace{-4mm}
Every occupation corpus, denoted by $o_n \in O$ can determine a distinguished occupation that is variously correlated with cognitive features. Given the collection of cognitive features, where each of them is associated with a single short-text content $t_j.P_q$ possessed by the author $a_m$, the inference model can estimate a relevant career for $a_m$.
\vspace{-6mm}
\subsection{Problem definition}
\label{prob:problem_def}
\vspace{-4mm}
\begin{problem}
	\label{prob:extract_linguistic_features}
	(extracting linguistic features)
	\textit{The linguistic features are hierarchically affected by cognitive-semantics. Given the short-text content $t_j$, our aim is to extract the related linguistic features $l_k$ to $t_j$.}
\end{problem}
\vspace{-4mm}
\begin{problem}
	\label{prob:mapping_linguistic_cues_to_cognitive_features}
	\textit{(mapping linguistic cues to cognitive features) The cognitive features are elucidated by linguistic features.  Given the linguistic cues $L_k$, our aim is to map $L_k$ to the pertinent set of cognitive features $P_q$.}
\end{problem}
\vspace{-4mm}
\begin{problem}
	\label{prob:inferring_professions}
	(inferring professions)
	\textit{Given the cognitive features $P_q$, our aims is to estimate an accurate occupation $o_n$ for $a_m$.}
\end{problem}
\vspace{-6mm}
\begin{figure}[htp]
	\tiny
	\centering
	\includegraphics[width=0.7\textwidth]{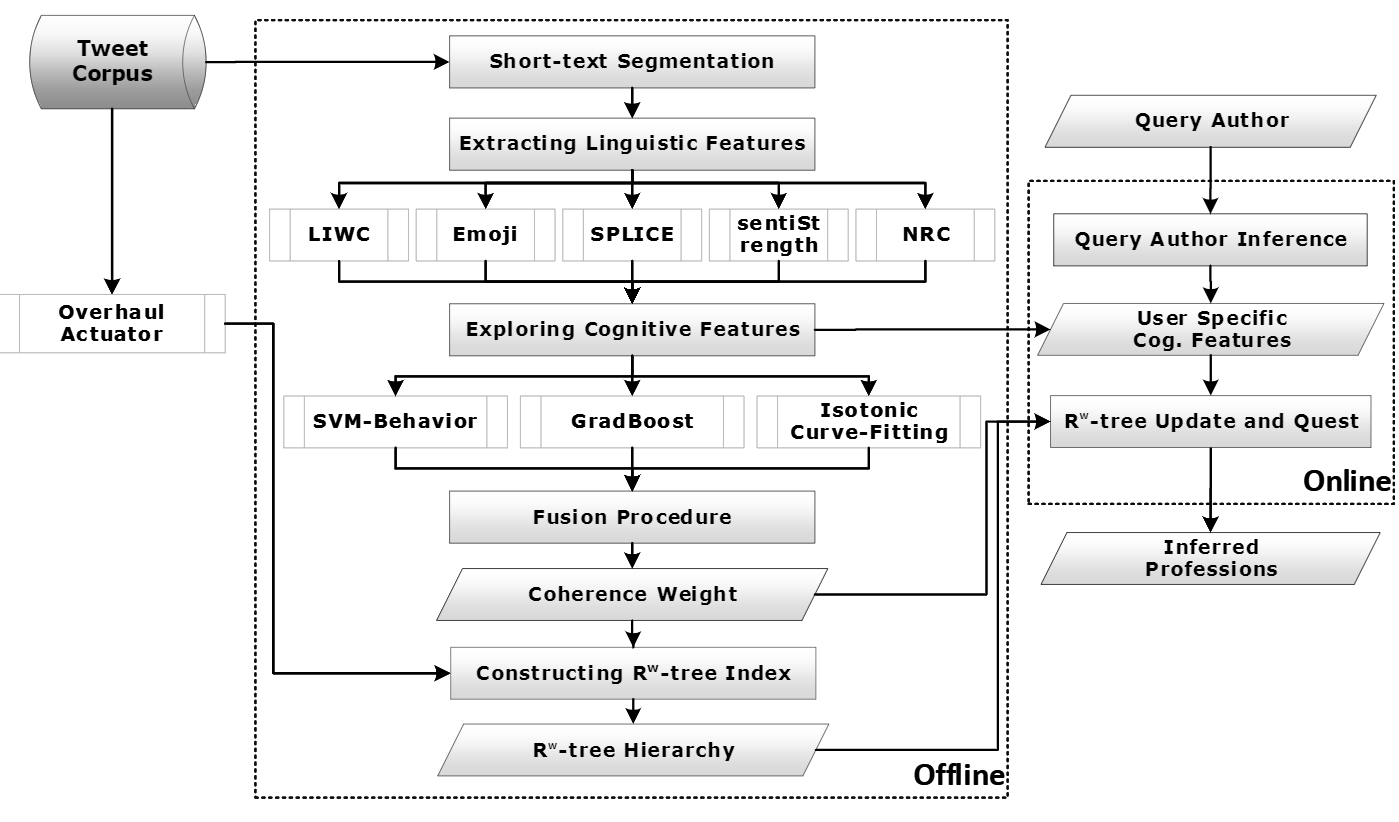}
	\vspace{-4mm}
	\caption{\small Framework}
	\label{fig:Framework}
	\vspace{-6mm}
\end{figure}
\vspace{-15mm}
\subsection{Framework Overview}
\label{Framework-Overview}
\vspace{-4mm}
The problem of inferring professions from short-text contents (Prob. \ref{prob:inferring_professions}) includes two primary steps: (1) to extract linguistic features $L_k$ from short-text $t_j$ (Prob. \ref{prob:extract_linguistic_features}). (2) to map the extracted linguistic feature $L_k$ to the most correlated cognitive cues $P_q$ (Prob. \ref{prob:mapping_linguistic_cues_to_cognitive_features}) that are useful to predict the jobs.
Figure \ref{fig:Framework} illustrates our proposed unified framework that can estimate the occupation of short-text authors through leveraging the cognitive features. In the offline part, we utilize the textual information of the query author to retrieve the cognitive orientation of an individual toward a specific occupation. 
From one side, we take advantage of segmentation in a Term Graph (TG) \citep{hua2015short} to understand coherence weight for each given term, and from the other side, we utilize the Pointwise Mutual Information to better infer the token-wise co-occurrences. Subsequently, we extract linguistic features that conjointly reveal the language habits and the latent semantic knowledge from author contents. We utilize cognitive-semantic algorithms such as LIWC, Emoji, SPLICE, sentiment Strength, and NRC lexical dictionaries to exploit cognitive features. We then map the linguistic features of each short-text $t_j.L_k$ to the pertinent cognitive features $t_j.P_q$. In response, aiming to estimate the professions, we fuse three of our behavior machine learning modules including SVM-Behavior, Gradboost behavior, and the curve fitting to calculate the joint coherence weight. In a nutshell, our proposed framework effectively acts as a triplet information pipeline. We collectively transform the cognitive features into trilateral modules including clustering, boosting, and curve fitting to identify professions, where the novel $R^w$-tree accommodates occupational boundaries. In the online part, we firstly compute contextual vectors for the query author and exploit pertinent cognitive features. To continue, we employ the update and quest procedures in the efficient $R^w$-tree to attain the top-k most relevant professions for the given query author.
\vspace{-5mm}
\section{Methodology}
\label{sec:methodology}
\vspace{-3mm}		
\subsection{Offline Processing}		
\label{met:offline_proc}		
\vspace{-3mm}
\subsubsection{Short-text Contents Segmentation}
\label{met:preproc}
\vspace{-4mm}
As described in Challenge 1 due to noise and grammatical errors (e.g. slang phrases and ambiguous words), common text mining approaches including tokenization \citep{tadesse2018personality} cannot fully obtain either the semantics. To address this issue, we take a three steps procedure: noise reduction, tokenization, and significance computation. For noise reduction, we improve the repeated characters \textit{(e.g. "woooow" to "wow")}, remove emotion irrelevant contents, and apply normalization to enrich the contents. While we use the slang knowledge-base\citep{wu2018slangsd} to replace the abbreviations \textit{(e.g. before instead of b4)}, we use word mover's distance \citep{beaver2020machine} to compute the expansion impact through word vectors.\\
Hua et al. \citep{hua2015short} associate a weight to each segment solely based on the term graph, treating each term as a node and neglecting the value of semantic measures. In response, as verbalized in Eq. \ref{formula:score_segmention}, we modified the associate score s(x, y) of each segment:
\vspace{-3mm}
\begin{equation}
\label{formula:score_segmention}
s(x, y) = \max_{\substack{i,j}}(\epsilon, \frac{w(x_i)+w( y_j)}{2}C(x_i, y_j))
\end{equation}
Here $\epsilon > 0$ designates a minimum positive score where $x_i$ and $y_j$ are the given terms and $C(x_i, y_j)$ reflects the semantic coherence between the words. we calculate in the offline phase. Also, $w(x_i)$ and $w(y_j)$ denote the weight of semantic-linguistic features. We calculate the edge score as the maximum score between the corresponding terms. Inherently, we distinguish between terms and phrases. A term is a single vocabulary with semantics and suitable for cognitive features \ref{met:exploring_cognitive_features}. 
Term frequency is the number of times a term appears in the contents of an individual author. Similarly, a phrase comprises two or more terms $h_p = \{e_t|t = 1, ..., i\}$ where the terms can not overlap with each other \textit{(i.e. $\forall t, e_t \cap e_{t+1} = \emptyset$)}. ``bright morning'' is a phrase containing two terms of ``bright'' and ``morning''. Accordingly, the corpus $C$ includes the short-text contents generated by all the authors. Correspondingly, we compute the term significance relying on the term-document matrix that is capable to alter Na\"ive word combinations with meaningful phrases. In our previous work \citep{hosseini2014location, hosseini2019leveraging} we empirically present that the Symmetric Conditional Probability (SCP) can exploit the phrases better than the Point-wise Mutual Information. Similarly, as Eq. \ref{formula:scp} shows we employ a modified version of the SCP to integrate the corresponding semantic score through term co-occurrence:
\vspace{-3mm}
\begin{equation}
\label{formula:scp}
SCP(x_i, y_j) = log \frac{s(x_i, y_j) Pr(x_i, y_j)^2}{\frac{1}{n-1}\sum_{i,j=1}^{n-1}Pr(x_i)Pr(y_j)}
\end{equation}
Here $x_i$ and $y_j$ denote the terms and $Pr()$ designates the probability of the phrase. As verbalized in Eq. \ref{formula:score_segmention}, the $s(x_i, y_j)$ computes the edge score for the segment. In an inverse term frequency manner, we observe that the higher the popularity of the phrase, the more authors will tend to utilize it. On the one hand we rank all possible segments by SCP, and on the other hand, we remove the terms and phrases that are excessively used by numerous authors. Through such arrangement, the SCP measure can eliminate uninformative phrases. Table \ref{tbl:locations} presents the notations.
\vspace{-8mm}
\begin{table}[h]
	\centering
	\def\arraystretch{1.5}
		\caption{Table of important symbols}
	\label{tbl:locations}
	\vspace{-3mm}
	\begin{tabular}{|l|l|l|l|}
		\hline
		\textbf{Definition}	&	\textbf{Notation}                                                                                                                                                                       
		\\ \hline
		\begin{tabular}[c]{@{}l@{}}\textit{sample short-text}\end{tabular} &
		\begin{tabular}[c]{@{}l@{}}{ $t_{j}$}\end{tabular} 
		\\ \hline	
		\begin{tabular}[c]{@{}l@{}}{\textit{The weight corresponding to the linguistic features $l_k$}}\end{tabular} &
		\begin{tabular}[c]{@{}l@{}}{$w_k$}\end{tabular} 
		\\ \hline	
		\begin{tabular}[c]{@{}l@{}}{\textit{The set of cognitive features for $t_j$}}\end{tabular} &
		\begin{tabular}[c]{@{}l@{}}{$t_j.P_q$}\end{tabular} 
		\\ \hline
		\begin{tabular}[c]{@{}l@{}}{\textit Cluster weight}\end{tabular} &
		\begin{tabular}[c]{@{}l@{}}{$\boldsymbol{\hat{w}_{cluster}}$}\end{tabular} 
		\\ \hline	
		\begin{tabular}[c]{@{}l@{}}{\textit Boosting score}\end{tabular} &
		\begin{tabular}[c]{@{}l@{}}{$\boldsymbol{\hat{w}_{boost}}$}\end{tabular} 
		\\ \hline
		\begin{tabular}[c]{@{}l@{}}{\textit Modified isotonic curve}\end{tabular} &
		\begin{tabular}[c]{@{}l@{}}{$\boldsymbol{\hat{w}_{curve}}$}\end{tabular} 
		\\ \hline
		\begin{tabular}[c]{@{}l@{}}{\textit Coherence weight}\end{tabular} &
		\begin{tabular}[c]{@{}l@{}}{$\hat{w}_C(P_i,o_j)$}\end{tabular} 
		\\ \hline
	\end{tabular}
	\vspace{-2mm}
\end{table}
\subsubsection{Exploring Cognitive Features}
\label{met:exploring_cognitive_features}
\begin{figure}[htp]
	\tiny
	\centering
	\includegraphics[width=0.7\textwidth]{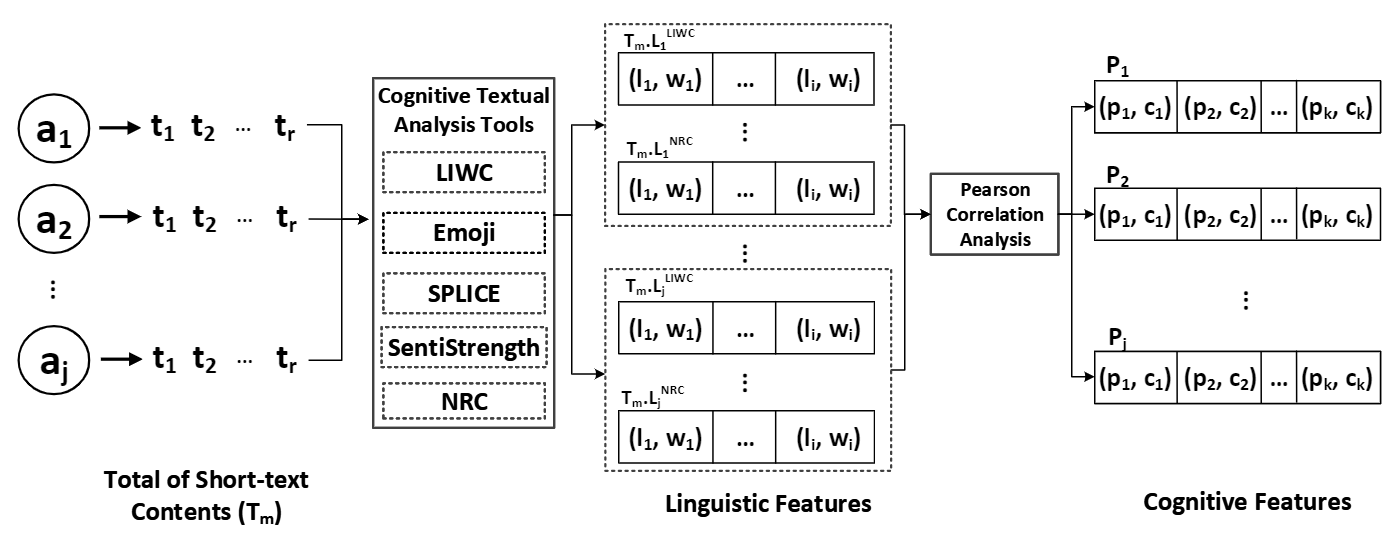}
	\vspace{-3mm}
	\caption{\small LESSN: Exploring Cognitive Features}
	\label{fig:exploring_cognitive_features}
	\vspace{-6mm}
\end{figure}

\begin{figure}[!htp]
	\tiny
	\centering
	\includegraphics[width=0.7\textwidth]{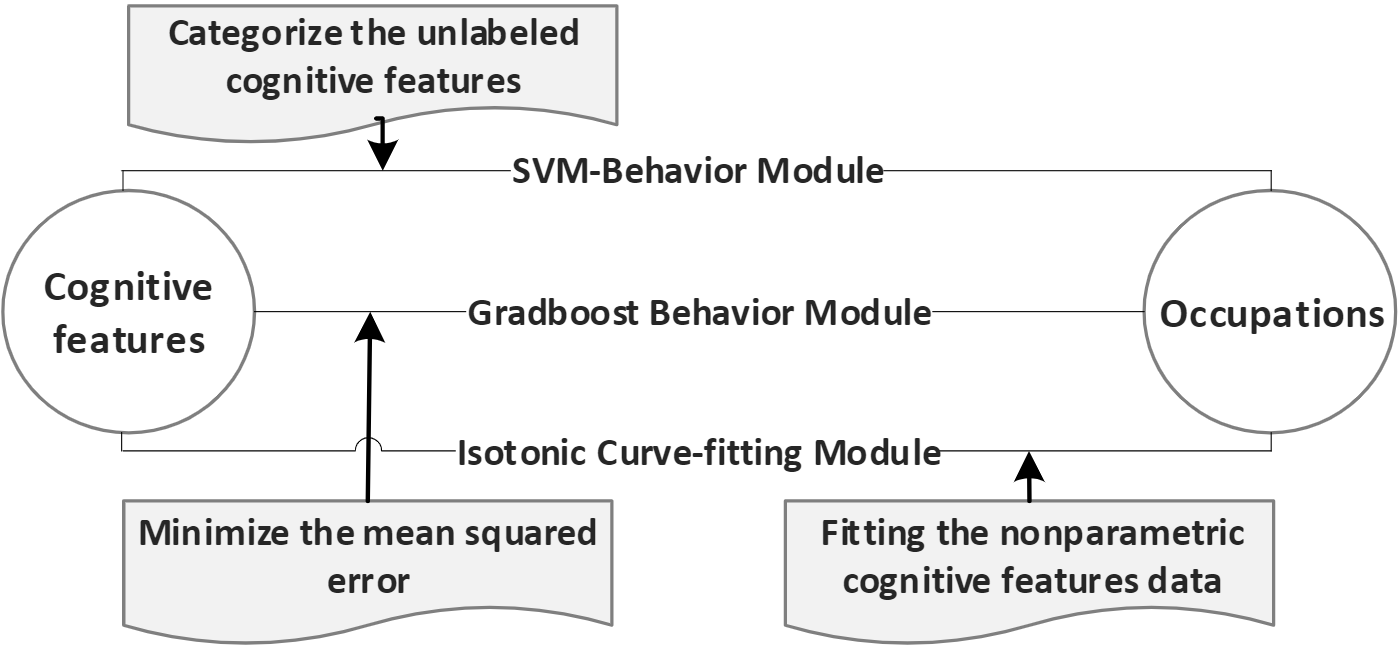}
	\vspace{-3mm}
	\caption{\small Weights vector for inferring professions}
	\label{fig:methodology_weight}
	\vspace{-7mm}
\end{figure}

\vspace{-4mm}
Inherently, we can apply the cognitive textual analysis tools on short-text vectors to retrieve the cognitive orientation of the authors. Hence, we argue that external knowledge (e.g. LIWC) \citep{ong2017personality} is indispensable to correctly retrieve the semantic of short-text vectors. To this end, we can extract linguistic features by using external knowledge of the short-text contents. Consequently, we can reveal the hidden relevance between cognitive and linguistic features. The genuine linguistic features turn latent in short-text contents. The current methods in extraction of linguistic features \citep{tadesse2018personality, ong2017personality} fail to attain the latent features from short-text vectors. To address this issue, as illustrated in Fig. \ref{fig:exploring_cognitive_features}, we propose a novel model, named as \textit{LESSN}, to disclose both linguistic and cognitive features. Given $T_m = \{t_1, t_2, ..., t_r\}$ as the set of short-text vectors corresponding to author $a_m$, the LESSN utilizes five complementary knowledge bases including LIWC, EMOJI, SPLICE, SentiStrength, and NRC to cover various aspects that we further equip them with the weights correlated to each of the linguistic features. Every given weight $w_k$ normalizes the linguistic feature frequency for $T_m$ using the same frequency from corpus $C$.\\
Linguistic Inquiry and Word Count (LIWC) knowledge-base \citep{tadesse2018personality} can extract the terms like $e_t \in t_j$ that are capable to reveal the  feelings, thinking styles, and social concerns that are particularly pertinent to the professions. Here $T_m.L_k^{LIWC} = \{k=1, ..., i|(l_k, w_k)\}$ represents the extracted features by LIWC where $w_k$ is the LIWC weight for a linguistic feature $l_k$. \\
\textit{EMOJI} \citep{giuntini2019feel} can constitute the attitude for emotion analysis and investigates the popular emojis for each cognitive feature based on the overall frequency $T_m.L_k^{EMOJI}  = \{k=1, ..., i|(l_k, w_k)\}$.\\ 
\textit{SPLICE} is another linguistic analysis mean \citep{tadesse2018personality} that we utilize to extract various features with sentiment poles to report self-evaluation results for the given author based on complexity and readability scores $T_m.L_k^{SPLICE}$.\\
\textit{SentiStrength} \citep{sitaraman2014inferring} is capable to estimate the strength sentiment positivity in informal short text contents, ranges from -1 (not negative) to -5 (extremely negative) and from 1 (not positive) to 5 (extremely positive). $T_m.L_k^{Senti} = \{(l_{neg}, w_{neg}), (l_{pos}, w_{pos})\}$ is the binary output of SentiStrength for the author $a_m$.\\
\textit{NRC} \citep{maheshwari2018revealing} is a lexicon that contains more than 14,000 distinct English words annotated with 8 emotions (anger, fear, anticipation, trust, surprise, sadness, joy, and disgust), and two sentiments (negative, positive). To achieve the features $T_m.L_k^{NRC}$ for the author $a_m$ we count the number of terms $e_t$ in each of emotions and sentiment poles. Successively, we can employ the simple but effective Pearson metric \citep{benesty2008importance} to measure the linear relationship between each weight of linguistic features $w_k$ and the impression scores of the cognitive features.
The Fig. \ref{fig:exploring_cognitive_features} illustrates how we attain the cognitive features, denoted by $P_q = \{q=1, ..., k|(p_q, c_q)\}$. Accordingly, each pair of $(p_q, c_q)$ represents a cognitive feature $p_q$ and its correlation range $c_q \in [-1, 1]$. While $c_q$ renders a linear relationship between $l_k$ and $p_q$, the values -1 and +1 indicate respective negative and positive absolutes and the value of zero $c_q=0$ exhibits irrelevance. 
Finally, as illustrated in Figure \ref{fig:methodology_weight}, we can invoke three edge weights such as, clustering \ref{met:svm}, boosting \ref{met:gradiantboost}, and the fitting \ref{met:curve_fitting} to collectively compute the association between cognitive features and occupations. 
Such portmanteau consisting of three properties can better characterize the correlations between the nodes in two categories of features and professions.
While the cluster module automatically groups the features based on the professional boundaries, the supervised boosting model estimates the relevance between cognition and occupation, and the fitting curve can track the variations in input features.
\vspace{-6mm}
\vspace{-1mm}
\begin{itemize}
	\item[$\bullet$] \textit{Cluster($\boldsymbol{\hat{w}_{cluster}}$)}: explains how the cluster pertaining each set of cognitive features $P_q$ is correlated with a specific occupation boundary $o_n$. To this end, we invoke support vectors of the SVM-behavior to categorize unlabeled cognitive features. The support-vector-clustering can subsequently form the clusters that expose occupational boundaries.
	\item[$\bullet$] \textit{Boost($\boldsymbol{\hat{w}_{boost}}$)}: optimizes a loss function based on mean squared error (MSE). In this model, each cognitive feature forms a binary tree, where the goal in the training procedure is to minimize the MSE ratio between the set of trees (a cognitive feature set) and each pertinent occupation.
	\item[$\bullet$] \textit{Curve($\boldsymbol{\hat{w}_{curve}}$)}: aligns every non-parametric set of cognitive features to a designated profession. We utilize the common but tractable Isotonic Curve-Fitting (ICF) module to pursue cognitive transformations.
\end{itemize}
\vspace{-6mm}
\vspace{-4mm}
\subsubsection{SVM-Behavior Module}
\label{met:svm}
\vspace{-4mm}
As elucidated in Section \ref{sec:intro}, the cognitive features can signify a specific profession. However, since the cognitive features are non-linear and multiplex, devising an effective mathematical model to infer pertinent professions is a tedious task. 
Inherently, it is neither easy to learn the prior knowledge about the occupation clusters nor feasible to estimate the number of such clusters solely based on the noisy short-texts. Hence, we empirically study the clustering procedure from these two perspectives. 
On the one hand, we utilize TFIDF \citep{dadgar2016novel} statistics to compute the significance of the non-stop terms and on the other hand, we combine the results with the set of pertinent cognitive features, denoted by $\boldsymbol{\hat{w}_{cluster}}$. 
Therefore, we argue that the cognitive-ware SVM-Behavior module can perceive the professions better than the support vector machine \citep{shafiabady2016using}. Also, we devise a jointly supervised classifier named as SVM-Behavior, which not only tracks the cognitive features but also consumes the short-text TFIDF vectors. Accordingly, to overcome the curse of dimensionality, we devise a Support Vector Cluster (SVC) module to improve the feature selection and decrease the number of dimensions. From one side, the SVC module projects the textual vectors to the higher dimensional space of the occupations and from the other side, it mutually asserts the Gaussian kernel to further incorporate cognitive orientations.\\
Given two inputs including the set of cognitive features and short-text TFIDF vectors, respectively denoted by as $\vec{P_i}$ and $\vec{d_j}$, we can present the corresponding profession-aware labels by $o_i$ and $o_j$. As clarified in Eq. \ref{formula:svm_kernel_function}, the training data consumed by the Gaussian kernel $k(\vec{P}_i, \vec{d}_j)$ can comprise both cognitive features and short-text TFIDF vectors. We can address Eq. \ref{formula:svm_kernel_function} using the quadratic algorithms.
\vspace{-3mm}
\begin{equation}
\label{formula:svm_kernel_function}
\begin{aligned}
&max \quad \sum_{i=1}^{l}\alpha_i - \frac{1}{2}\sum_{i=1}^{l}\sum_{j=1}^{l}\alpha_i \alpha_j o_i o_j k(\vec{P_i},\vec{d_j} ) \\
&s.t \quad \sum_{i=1}^{l}\alpha_i o_i = 0, \quad  0 \leq \alpha_i \leq C, \quad i = 1, 2, ..., l\\
\end{aligned}
\end{equation}
Nevertheless, we need to accommodate the optimized threshold of $C$ inside the occupational boundary to attain a smaller-marginal hyperplane. Moreover, we utilize the parameters $\alpha_i$ and $\alpha_j$, corresponding to the Lagrange multipliers of $\vec{P_i}$ and $\vec{d_j}$ to retrieve the maximum and minimum local boundaries. As Eq. \ref{formula:svm_weight_feature} shows, $k_F$ denotes the cognitive Gaussian kernel function with $F$ as the cognitive linear matrix for the given input space and $\eta$ as the bias parameter to ensure the symmetry. 
\vspace{-3mm}
\begin{equation}
\label{formula:svm_weight_feature}
\begin{aligned}
&k_F(\vec{P_i}, \vec{d_j}) = exp \Big(-\eta\parallel \vec{P_i}^TF - \vec{d_j}^TF \parallel^2\Big) \\
& \hspace{15.5mm}   = exp \Big(-\eta\big((\vec{P_i} - \vec{d_j})^T FF^T (\vec{P_i} - \vec{d_j})\big)\Big)
\end{aligned}
\end{equation}
For the next stride, the SVC module takes the Gaussian kernel $k_F$ to project the input data points to the higher dimensional cognitive feature space. To promote the efficiency of the SVC model, we further adjust the boundary information of occupations based on the cognitive features. This proves that our supervised classifier can concurrently support both flexibility and scalability. The classifier module is capable to collect the appropriate boundaries and at the same time can reduce redundant noise. Subsequently, the SVC module can retrieve the optimized sphere, ensuring a minimum radius that can comprise the majority of the projected data samples. Such sphere when mapped back to the data space, can be partitioned into several components, each exposing an isolated cluster of instances. Eq. \ref{formula:svm_svc} reflects the constraint on the spherical radius, denoted by $R$:
\vspace{-3mm}
\begin{equation}
\label{formula:svm_svc}
\begin{aligned}
&\min_{\substack{R, \lambda, \phi}}\dfrac{1}{2} R^2 + C \sum_{i}^l \xi_i\\
& s.t. \quad \parallel\phi(P_i) - \lambda \parallel^2 \leq R^2 + \xi_i\\
& \lambda = \sum_{j}^l \alpha_j\phi (d_j)
\end{aligned}
\end{equation}
Here, $\xi_i \geq 0$ optimizes the empty boundary that relaxes the strict condition of non-linear separability, so that each input data can be observed inside the occupation boundary. We adjust the threshold $C$ to control the penalty of the noise in input data. While $\lambda$ is the spherical-center of each occupation boundary, we map $\phi(P_i)$ and $\phi(d_j)$ as a non-linear transformation from non-separable input data to the linear space to classify the features corresponding to professions. As Eq. \ref{formula:svm} shows, we compute the clustering weights based on the training data, where the weight $\boldsymbol{\hat{w}_{cluster}}$ computed by the SVM-behavior can assess the selection procedure of the cognitive features.
\vspace{-4mm}
\begin{equation}
\label{formula:svm}
\begin{aligned}
&\boldsymbol{\hat{w}_{cluster}} = \sum_j^l a_j o_j - \sum_{i,j}^l a_i a_j o_i o_j k_F(\vec{P_i}, \vec{d_j}) \\
\end{aligned}
\end{equation}
Inherently, it is neither easy to learn prior knowledge about the occupation clusters nor feasible to estimate the number of such clusters solely based on the noisy brief contents. To this end, the SVC model adjusts the boundary information of occupations and employs a supervised classifier to reduce redundancy and noise in short-text vectors.
\vspace{-4mm}
\subsubsection{Gradboost Behavior Module}
\label{met:gradiantboost}
\begin{figure}[!htbp]
	\tiny
	\centering
	\includegraphics[width=0.6\textwidth]{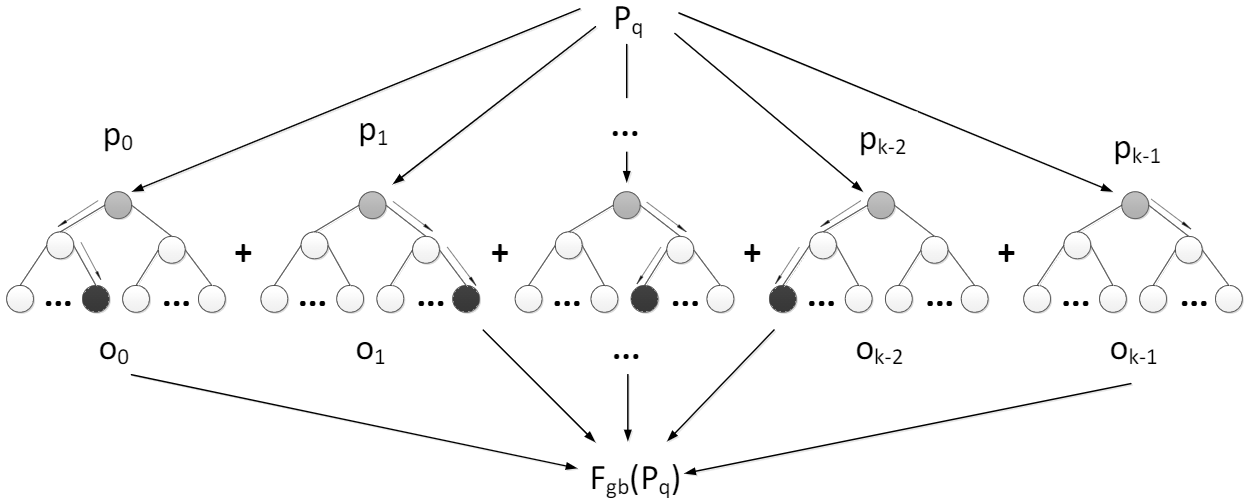}
	\vspace{-3mm}
	\caption{\small Boosting Behavior Tree}
	\label{fig:BBT}
	\vspace{-6mm}
\end{figure}

\vspace{-4mm}
As explained in Section \ref{intro:challenge_inferring_module}, we devise a supervised boosting approach to iteratively improve the profession-related classifiers, where we increase the cognitive weights of the observations that are difficult to associate with a particular occupation and conversely reduce the weights for those that are easily correlated with a particular job. In other words, we employ the gradient boosting module to minimize the mean squared error between the cognitive features and occupations, denoted by $\boldsymbol{\hat{w}_{boost}}$. This module forms an ensemble of $K$ decision trees where each tree, so-called as Boosting Behavior Tree (BBT), is tightly correlated with a cognitive vector, subsequently used to estimate a target label. As illustrated in Figure \ref{fig:BBT}, $P_q$ denotes the set of exploited cognitive features and $o_q$ (e.g. $o_1$) represents the estimated occupation for the particular cognitive feature (e.g. $p_1$). Each BBT learns a distinct cognitive attribute and identifies an output gradient space $o_q$ for the input feature.
\vspace{-6mm}
Given the correlation metric between linguistic features and the specific cognitive feature, e.g. $p_0$, we divide each cognitive node (in gray) to a positive correlation ($c_q \geq 0$) on the left and a negative correlation ($c_q < 0$) on the right. The BBT model successively exploits a suitable occupation (black node) from the set of available jobs and divides the original cognitive feature space $R^{d}$ into occupational disjoint areas (leaves) where we assign a constant value $\gamma$ for each region. We compute $o_0$ by utilizing the loss function $\mathcal{L}(g_{i}, \gamma)$ that aims to minimize the empirical risk. While $g_i$ delineates the estimated occupation (black node) in Eq. \ref{formula:gboost}, we initialize the supervised learning by a constant value as $\gamma$.
\vspace{-4mm}
\begin{equation}
\label{formula:gboost}
o_{0} = argmin\sum ^{n} _{i=1} \mathcal{L}(g_{i}, \gamma) 
\vspace{-2mm}
\end{equation}
Eq. \ref{formula:gboost_gamma} and \ref{formula:gboost_F} updates the approximation of the currently estimated job $o_q^{R_l}$ using the output of the previous tree $o_{q-1}$. Since each occupation region corresponds to several cognitive features, we select $o_q^{R_l}$ as the most correlated dimension.
\vspace{-3mm}
\begin{equation}
\label{formula:gboost_gamma}
o_q^{R_l} = argmin\sum_{P_q}\mathcal{L}(g_q, o_{q-1} + h_q)
\end{equation}
\vspace{-3mm}
\begin{equation}
\label{formula:gboost_F}
o_q = o_{q-1} + h_q . o_q^{R_l}
\end{equation}
As formulated in Eq. \ref{formula:gboost_h}, $h_q$ signifies gradient statistic measure for each leaf node. To this end, we employ an iterative procedure where at each round we use a new cognitive feature tree to associate the input $p_q$ to the $L$-disjoint output region $\{R_{l}\}_l^L$. Here, $O_{m}$ represents the number of leaf nodes.
\vspace{-3mm}
\begin{equation}
\label{formula:gboost_h}
h_q = \sum_{l=1}^{O_m}g_{l} R_l
\end{equation}
The BBT method approximates the function $F_{gb}$ as an additive expansion of the cognitive feature learners ($o_i \in \{o_0, o_1, ..., o_{k-1}\}$). We can subsequently determine the best fitted occupation by applying the simple but effective majority voting on the learning outcomes, associated to all of the input feature trees. The boost score $\boldsymbol{\hat{w}_{boost}}$ elucidate the level of minimized MSE between input cognitive features and the output inferred occupation. For ease of implementation, we first calculate the $w_q$ weight for every input tree $p_q$ (Eq. \ref{formula:boost_weight_q}) and in sequence compute the boosting weight by Eq. \ref{formula:boost_weight}:
\vspace{-3mm}
\begin{equation}
\label{formula:boost_weight_q}
w_q = |o_q|(2-|h_q|)
\end{equation}
\vspace{-3mm}
\begin{equation}
\label{formula:boost_weight}
\boldsymbol{\hat{w}_{boost}} = \frac{1}{|K|}\sum^{k}_{q=1} w_q 
\end{equation}
It is notoriously difficult to learn the weak points to minimize the MSE measure. Hence, the BBT model invokes the cognitive trees to compute behavior functions individually, commencing with equal weights for all cognitive features.
\vspace{-3mm}
\subsubsection{Isotonic Curve-fitting (ICF) Module}
\label{met:curve_fitting}
\vspace{-4mm}
As elucidated in Section \ref{sec:intro}, many of cognitive features excessively single-out multiple occupations, instead of one. The problem can turn-up even more challenging where the cognitive features can evolve continuously. Hence, the solution is concerned with a set of non-parametric features with scattered distributions. Therefore, we are required to devise a novel module to direct features toward exclusive professions and simultaneously track the feature-specific variations. Since the cognitive features are inherently nonlinear, the linear regression approaches \citep{cui2018effect, tadesse2018personality} can not address the challenge. Conversely, we opt for an adopted Isotonic Curve Fitting model, dubbed ICF, that can equip our proposed framework with an unbiased fit to the input cognitive figures. We note that the ICF can successfully form a free-shaping line to follow the feature variations and get aligned toward the real observed data, that is not included in the capabilities of the polynomial regressors \citep{weidmann2017concurrent}. However, this approach can encounter computational and statistical overfitting issues in higher dimensions.
To address this concern, we build a curve weight between cognitive features data input and occupations $\boldsymbol{\hat{w}_{curve}}$ that can eventually converge to the global isotonic curve.
While the proposed modification can often gain better accuracy, it can indirectly resolve the data complexity challenge in the weights of the curve vectors.\\
As formulated in Eq. \ref{formula:isotonic}, given the numerical analysis of the correlation weights for the cognitive features, denoted by $P_i$, $f_i: [0,1]^d \rightarrow R^d$, our proposed sequential ICF module can retrieve a weighted least-squares fit $P_i \in R^d$ for the corresponding data-oriented vector $\alpha_i \in R^d$ that comes with the weight vector of $w_i \in R^d$.
\vspace{-4mm}
\begin{equation} 
\label{formula:isotonic}
min \sum_{i=1}^{k} w_{i}(P_{i} - \alpha_i)^2 \quad \textrm{where}  \quad P_{i} \leq P_{i+1} \quad \textrm{for}  \quad i = 1,2,...,k
\end{equation}
Given the observed data $f_i = \{i = 1, 2, ..., k|(P_i, o_i)\}$, $o_i$ can estimate the pertinent occupation for $P_i$ using Eq. \ref{formula:isotonic_y}:
\vspace{-3mm}
\begin{equation} 
\label{formula:isotonic_y}
o_i = f_0(P_i) + \mu_{\{i,j\}} \quad \textrm{where}  \quad i \geq 2
\end{equation}
Here, $f_0: [0,1]^d \rightarrow R^d$ is the Borel measure \citep{han2019isotonic} and $\mu_{\{i,j\}}$ denotes the independent noise, the dissimilarity ratio between the occupation and the cognitive features. The more the occupations are specified by the cognitive features, the less noise will be remarked. Due to excessive noise, we perceive unobserved heterogeneity in occupations that increases the fitting variations and affects generalizability. We bound the noise for the features using the mean and the variance that are respectively denoted by $\mathbb{E}[\mu_{\{i,j\}}]$ and $\sigma^2$: 
\vspace{-3mm}
\begin{equation} 
\label{formula:isotonic_noise}
\quad Var[\mu_{\{i,j\}}] = \sigma^2, \quad \forall i = P_q \quad \text{and} \quad j = o_n
\vspace{-2mm}
\end{equation}
Hence, we utilize the curve weight vectors $\boldsymbol{\hat{w}_{curve}}$ associated with every edge $(P_i, o_i)$ to minimize the uncertainty in the estimation procedure, caused by the noise. As Eq. \ref{formula:isotonic} shows, $o_i \in \{ o_1, ..., o_k\}$ observes a single instance out of $k$, corresponding to the weight $w_i \in \{w_1, ..., w_k\}$. We can then estimate the value for each occupation using the relevant cognitive features that are collectively approximated by the combinational variations (Eq. \ref{formula:isotonic_weight_vector}):
\vspace{-4mm}
\begin{equation} 
\label{formula:isotonic_weight_vector}
\boldsymbol{\hat{w}_{curve}} \propto \frac{1}{\hat{\sigma}_i^2} = \bigg[\sum_{i=1}^{k}(o_{i} - \bar{o_i})^2\bigg]^{-1}
\vspace{-2mm}
\end{equation}
While $\bar{o_i}$ is the mean of the occupation responses for the $i$th cognitive features, $o_i$ denotes the estimated occupation by variable levels.
\vspace{-4mm}
\subsection{Online Processing}	
\label{met:online_proc}	
\vspace{-4mm}
In the online phase, we estimate the occupation of the input query author using short-text contents through two tasks: (i) \textit{Query Author Inference} that collectively processes the contents of the input user and extracts the cognitive features. (ii) \textit{$R^w$-tree Update and Quest} that firstly refreshes the index of the $R^w$-tree by the coherence weight that we calculate using the user-specific features in the offline phase. Secondly, given the cognitive features of the input author, we unveil the most relevant occupations.		
\vspace{-4mm}
\subsubsection{Query Author Inference}
\label{met:including_query_author}
\vspace{-4mm}
The procedure of Query Author Inference comprises two main tasks: \textit{generating query author segments} and \textit{extracting cognitive features}. To generate the textual segments from query author contents, we firstly combine all short-texts into a single document. The reason is that the model is already built-in the offline section and the cold-start users that come with limited content can specifically benefit from such merged contentment. We note that posting a few short-text contents can not sufficiently reveal an adequate number of cognitive features to estimate the occupation of a cold-start user. To continue, we carry out noise reduction and tokenization. Similar to Sec. \ref{met:preproc}, we utilize the SCP model to extract final segments in the form of output N-Grams, resulting in sufficient semantics.
Aiming to explore the cognitive features, we adopt the proposed approach in Sec. \ref{met:exploring_cognitive_features} that can be done fairly quickly for the single query author. We initially exploit the latent linguistic features by textual analysis tools and the external knowledge-base. Subsequently, the LESSN module can attain cognitive cues. On the one hand, we integrate all exploited linguistic features as an output set, and on the other hand, the Pearson correlation can retrieve the relationship between linguistic and cognitive features.\\
\textit{What is the usage of the Overhaul Actuator?} Inherently, the more varied the exploited linguistic features are, the higher the diversity of the cognitive features will be. The overhaul actuator is triggered by the sense where the level of variations in linguistics passes a threshold and the framework gets alarmed to rearrange the features within $R^w$-tree. Such an automatic procedure, not only improves the correlation between linguistic and cognitive features, but also improves the generalizability of our proposed model through assigning more accurate professions to the unseen forthcoming authors.
\vspace{-4mm}
\subsubsection{Computing Coherence Weight}
\label{met:coherence_weight}
\vspace{-4mm}
In this section, we aim to adjust the parameters that can collectively explain the correlation between a set of cognitive features and a single occupation.
Inherently, the professions follow a hierarchical pattern. For instance, a pair of occupations including C\# and C++ programmers firstly inherit from computer engineering. Accordingly, a computer engineer can comprise the cognitive features that belong to both profession leaves: C\# and C++ programmers. Adversely, if an individual possesses the cognitive features of a C\# programmer, he can be categorized as a computer engineer too. To model the hierarchy, we utilized $R^w$-tree as a modified version of $R$-tree \citep{balasubramanian2012state}. While the $R^w$-tree structure is consistent with the agglomerative attribute of occupations, the tree blocks can accommodate the set of cognitive features for each specific occupation.\\
As verbalized in Eq. \ref{formula:coherence_weight}, the coherence weight, denoted by $\hat{w}_C(P_i,o_j)$ can collectively associate a set of cognitive features $P_i$ to an occupation boundary $o_j$. Based on Sec. \ref{met:exploring_cognitive_features} we can compute the correlations through $\hat{w}_{cluster}$, ($\hat{w}_{boost}$), and $\hat{w}_{curve}$. Instead of na\"ive tuning, we adopt the simple but effective PageRank algorithm to collectively learn the best values for the bias parameters, $\alpha$, and $\beta$.
\vspace{-3mm}
\begin{equation} 
\label{formula:coherence_weight}
\hat{w}_C(P_i,o_j) = (1-\alpha-\beta) \times \hat{w}_{cluster} + \alpha \times \hat{w}_{boost} + \beta \times \hat{w}_{curve}
\vspace{-1mm}
\end{equation}
\textbf{Lemma.} \textit{Given an arbitrary number of classes, one can exploit the occupation boundaries via cognitive orientations in an iterative adjustment procedure on $R^w$-tree.}\\
\textit{Proof:} Let the number of arbitrary classes be $\kappa$. Hence, there will be a unified final result for constructing $\kappa$ classes. Based on the coherence weight adjustment $\hat{w}_C^\kappa(P_i,o_j)$ on the trilateral scores each result can represent a particular cognitive feature for each of given $\kappa$ classes. Thus, there will be $|\kappa|$ decisions associated with each of modules $\hat{w}_{cluster}^\kappa$, $\hat{w}_{boost}^\kappa$, and $\hat{w}_{curve}^\kappa$, where we can collectively maximize the accuracy of the final estimated class in the $R^w$-tree structure. Consequently, using a repetitive adjustment procedure, we can take advantage of the decision on the $R^w$-tree structure, commencing with the first score denoted by $\hat{w}_C^1$ on $\kappa = 1$. With this logic, the coherence weight can construct the $R^w$-tree hierarchy on an arbitrary set of occupations.
\vspace{-7mm}
\begin{figure}[!htbp]
	\tiny
	\centering
	\includegraphics[width=0.6\textwidth]{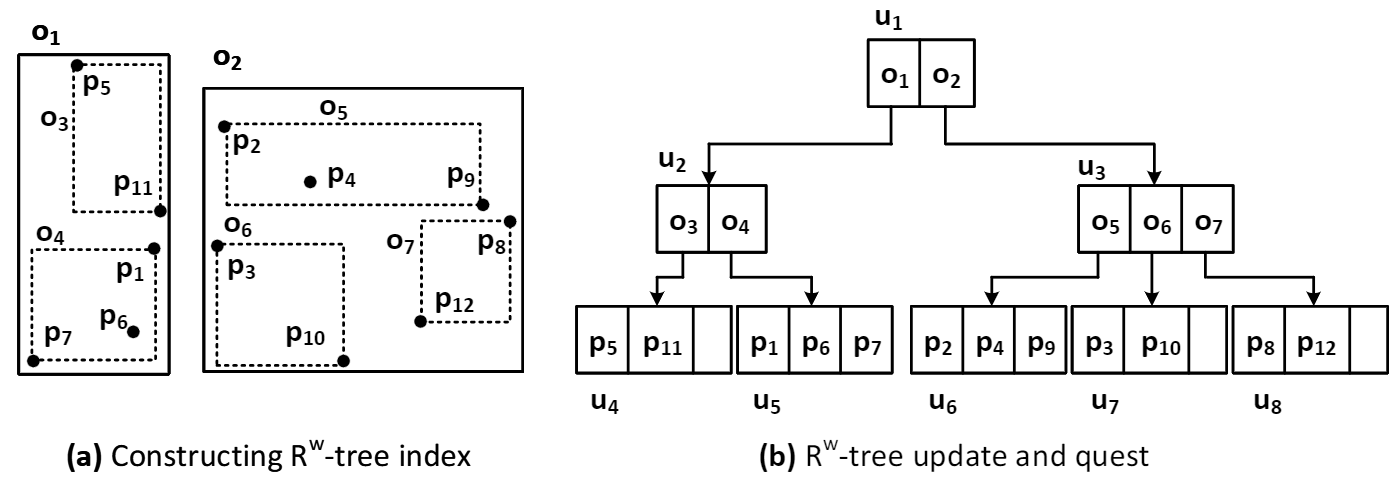}
	\vspace{-3mm}
	\caption{\small $R^w$-tree data structure to gain the occupational rectangle}
	\label{fig:r-tree}
	\vspace{-8mm}
\end{figure}
\vspace{-4mm}
\subsubsection{Constructing $R^w$-tree Index Hierarchy}
\label{met:r_tree}
\vspace{-4mm}
We employ a modified version of the $R$-tree \citep{balasubramanian2012state} named as $R^w$-tree and utilize collection weight $\hat{w}_C(P_i,o_j)$ to distinguish the relevant boundaries. The $R^w$-tree aims to find a profession based on a small set of cognitive feature nodes.
As Figure \ref{fig:r-tree} illustrates, each $p_q$ denotes a cognitive orientation point, where $\delta$ highlights the minimum number of acceptable points in each profession-related rectangle. For instance in Figure \ref{fig:r-tree} (a), where the $\delta$ threshold is set to $3$, we can nominate the $o_4$ profession boundary, comprising $p_1$, $p_6$, and $p_7$. In other words, we aim to specify $\delta$ number of cognitive points to create an occupation rectangle.
We consider a parameter $u_i$ for nodes. The internal $u_i$ keeps the outcome of the Minimum Bounding Rectangle (MBR)\citep{yuan2016algorithm} for the number of cognitive features.\\
As algorithm \ref{algorithm:r-tree} shows the quest and update methods to obtain the list of occupations $O_{result}$ based on the input cognitive features $r$ in our $R^w$-tree, denoted by $\hat{R}^w$. Where $r$ represents the input cognitive orientations, $O_{result}$ highlights the output list of occupations. Moreover, we use $u$ as the bounding box for the $R^w$-tree to appoint a cognitive orientation node for each leaf and the occupation node for the parent nodes. \\
To start, we achieve the list of cognitive features in the bounding box $O_{rec}$ by initializing the rectangles in the $R^w$-tree(line 2).
We use the update and quest procedures to find the suitable occupational boundaries for the input cognitive feature set of $r$. If the input is not discovered in $O_{rec}$, we trig the update procedure through the $MBR$ function (line 12) and update the occupational rectangles based on the $r$ and $\delta$ threshold. On the contrary, where $r$ exists in the $O_{rec}$, the quest method (line 8) returns the occupational boundaries that are pertinent to the input cognitive feature set.
Where we assert that the node $u$ comes with a child node, we recursively call the quest procedure to reveal all the children in the hierarchy (line 23).
Otherwise, where the node $u$ does not have a child (as an occupation rectangle), the search procedure can compare the input $r$ with the properties in the current object $u.P$ (line 27). Eventually, if the resultant output is not empty, the algorithm will return the parent of the node $u$ that represents the profession rectangle (lines 8-9), finalizing in a list of occupations $O_{result}$.\\
\begin{algorithm}[htbp]
	\caption{$R^w$-tree update and quest procedures}
	\label{algorithm:r-tree}
	\textbf{Input:} $\hat{R}^w, r, \delta$\\
	\textbf{Output:} $O_{result}$
	\begin{algorithmic}[1]
		\STATE $O_{result} \leftarrow []$
		\STATE $O_{rec} \leftarrow initializeRectangle(\hat{R}^w)$
		\IF{$r$ $!\in$ $O_{rec}$}
		\STATE $\hat{R}^w \leftarrow Update(r)$
		\STATE $O_{rec} \leftarrow initializeRectangle(\hat{R}^w)$
		\ENDIF
		\FOR{$o$ in $\hat{R}^w.nodes()$}
		\STATE $Quest(o,r)$
		\ENDFOR
		\textbf{\STATE Procedure Update($r_u$)}:
		\IF{$r_{u}.count()$ $\geq$ $\delta$}
		\STATE $u^\prime$ $\leftarrow$ $MBR(\hat{R}^w, r_{u})$
		\STATE \textbf{return} $u^\prime$			
		\ELSE
		\STATE print('Minimum $\delta$ orientations are required.')
		\STATE \textbf{break}
		\ENDIF	
		\textbf{\STATE Procedure Quest($(u_q, r_q)$)}:
		\IF{$u_q.child()$}
		\FOR{$v$ in $u_q.children()$}
		\STATE $Quest(v,r_q)$
		\ENDFOR		
		\ELSE
		\IF{$r_q == u_q.P$}
		\STATE 	$o^\prime \leftarrow u_q.parent()$
		\STATE 	$O_{result}.append(o^\prime)$
		\ENDIF
		\ENDIF
		\STATE \textbf{return} $O_{result}$
	\end{algorithmic}
\end{algorithm}
\textbf{Overhaul Actuator.} 
Linguistic features may evolve during streaming. Accordingly, the distribution of the cognitive features associated with each of the career boundaries may significantly change in the $R^w$-tree. Inherently, such skews can affect the dependencies between cognitive features and negatively affect the performance of the profession inference module. Hence the Overhaul Actuator should estimate under what intervals one must re-initiate the $R^w$-tree. To this end, we employ an adapted divide and expansion method to approximately estimate the actuation interims. As algorithm \ref{algorithm:trigger} shows, we first take the whole or a portion of the data $\mathcal{D}$ if the size is massive. We then divide the dataset into two equal parts denoted by $d = \{d[0], d[1]\}$. To commence the evaluation part, we process the first subset $d[0]$ to explore the cognitive features and acquire the occupational boundaries, based on which we can employ any model $Sub_{tree}()$ to exploit the set of sub-trees from $\tilde{R}$. The evaluation function $E_{w}(\tilde{R})$ can measure the effectiveness $\omega^0$ of the novel job inference module using the proposed benchmark in Section \ref{benchmark}.
Subsequently, we incrementally add up the initial portion $d[0]$ by $\zeta$ percent from the second half $d[1]$. We then recompute the effectiveness of $\omega^i$ for iteration $i$. We can use the insufficiency ratio $(\omega^0 - \omega^i)$ to compute the error level in $R^w$-tree that is owing to the augmented data. Where the MAE size is greater than the threshold $\epsilon$, the sensing procedure can trig the actuation unit to enforce the reconstruction of the modified $R^w$-tree. Conversely, if the error rate does not exceed the $\epsilon$, one can justify that either the expansion rate has been selected incorrectly, resulting in overfitting, or the error is negligible and the next iteration should be followed.
\vspace{-4mm}
\subsubsection{Complexity Comparison: $R$-tree versus $R^{w}$-tree}
\label{Complexity}
\vspace{-4mm}
This section compares the ordinary $R$-tree against our proposed $R^{w}$-tree from a theoretical computer science perspective. Intuitively, every basic $R$-tree can manipulate the indexing rectangles, denoted by $o_j$, where each of them can specify an optimum number of pertinent cognitive features $P_i$. Note that our proposed $R^w$-tree structure dedicates a weight ($\hat{w}_C(P_i, o_j)$) to each $o_j$, initialized when the occupation node is inserted. Here is where the $R^w$-tree behaves differently compared to the trivial $R$-tree. Let $\gamma$ be the set of the cognitive features ($\mathbb{P}$) and $\iota$ as the number of boundary classes ($O$), in this case the original $R$-tree will run in $O(log^{\gamma + \iota}_{\iota})$. As elucidated in Section \ref{met:r_tree}, our proposed method constructs the $R^w$-tree that further divides the data into three levels based on the coherence weights. As algorithm \ref{algorithm:r-tree_insert} shows, we can assign the highest weight of occupations in the middle-level of $R^w$-tree and reserve the lowest weight for the top-level. Therefore, the expected time for $R^w$-tree to run each procedure will result in $O(log^{\gamma + \frac{\iota}{2}}_{\iota})$, with an apparent lower complexity.
\vspace{-6mm}
\begin{algorithm}[htbp]
	\caption{$R^w$-tree Overhaul Actuator Interval Estimation}
	\label{algorithm:trigger}
	\textbf{Input:} $\mathcal{D}, \zeta, \epsilon$\\
	\textbf{Output:} $|d[1]|$
	\begin{algorithmic}[1]
		\STATE $d \leftarrow Split(\mathcal{D}, 2)$
		\STATE $\tilde{R}\leftarrow Sub_{tree}(d[0])$
		\STATE $\omega^0$ $\leftarrow$ $E_{w}(\tilde{R})$
		\STATE $q_s \leftarrow d[0]$
		\FOR{$i$ in Range($1, |d[1]|/\zeta$)}
		\STATE $q_s \leftarrow q_s$ + Subset($d[1], i$)
		\STATE $\tilde{q}_s \leftarrow Sub(q_s)$
		\STATE $\omega^i \leftarrow \textbf{E}_w(\tilde{q}_s)$
		\IF{$\omega^0 - \omega^i \geq \epsilon$}
		\STATE \textbf{return} $i\times\zeta$
		\ENDIF
		\ENDFOR
		\STATE \textbf{return} $|d[1]|$
	\end{algorithmic}
\end{algorithm}
\vspace{-12mm}
\begin{algorithm}[htbp]
	\caption{Insertion in $R^w$-tree}
	\label{algorithm:r-tree_insert}
	\textbf{Input:} $\mathcal{O}_c, l, \hat{R}^w$\\
	\textbf{Output:} $\hat{R}^w$
	\begin{algorithmic}[1]
		\STATE $l \leftarrow l_i$
		\textbf{\STATE Procedure Insert($\mathcal{O}_c$)}:
		\STATE $r \leftarrow $ new $\hat{R}^w.Subtree()$
		\IF {$\mathcal{O}_c.child()$}
		\FOR{$i$ in $\mathcal{O}_c.children()$}
		\STATE $r^l \leftarrow \{i[orients]\}$
		\STATE $r^{l-1} \leftarrow \{i[name], i[weight]\}$
		\ENDFOR
		\ENDIF
		\STATE $r^{l-2} \leftarrow \{\mathcal{O}_c[name], \mathcal{O}_c[weight]\}$
		\STATE \textbf{return} $\hat{R}^w$
	\end{algorithmic}
\end{algorithm}\\
Based on algorithm \ref{algorithm:r-tree_insert}, each $\mathcal{O}_c$ node represents an occupation and comes with the weight, cognitive orient (the value of $\gamma$), and a name. The $l$ parameter specifies the number of levels in the tree. In the insertion procedure, we create a new sub-tree of occupation nodes for each boundary denoted by $r$, and examine the child nodes (line 4). Additionally, we store the cognitive features of the child node in the last level $r^l$, the child node of the boundary in the middle-level $r^{(l-1)}$, and the parent node of the boundary in the first level of the object $r^{(l-2)}$. Finally, we return the $R^w$-tree. Through applying the $R^w$-tree algorithms \ref{algorithm:r-tree} and \ref{algorithm:r-tree_insert} we can increase the performance of the procedures over the $R$-tree algorithm. To get the top-k nodes we consider the points that the middle-level boundary will gain a higher rank than the top-level. Hence the quest procedure will be divided ($\frac{\iota}{2}$) based on the values in the middle-level.\\
\textbf{Scalability.} where the size of the dataset is small, i.e. $|\gamma + \iota|$ = $10k$, the performance of both trees will be reasonable. However, in big data scenarios, where the number of boundaries augments, the performance of the $R$-tree gets flatten. But since $R^w$-tree accommodates the nodes based on the value of the weights in various levels, higher levels with lower weights, and vice versa, the performance of $R^w$-tree decreases less. We also observe that compared to the total number of boundaries in $R$-tree, the proposed $R^w$-tree can execute the construct, updates, and quest procedures based on the levels that are determined by the range of the weights.  
\vspace{-6mm}
\section{Experiments}
\label{sec:Experiment}
\vspace{-4mm}
We conducted extensive experiments on a real-world Twitter dataset \citep{hosseini2014location} to examine the performance of our cognitive-semantic approach in inferring of the professions from short-text contents. We used a machine with 64GB of RAM with Intel Core i7-7700K CPU of 4.20GHz. (Code and data)\footnote{https://sites.google.com/view/infer-jobs-cognitive-approach}.
\vspace{-6mm}
\subsection{Data}
\label{data}
\vspace{-4mm}
We collected the Twitter dataset \citep{hua2016understand} with more than 2 million tweets through \textit{Twitter API} and selected $5K$ users and subsequently retrieved up to 300 records from each tweet history. The final dataset contains $76K$ words of $16M$ collocations.
\vspace{-5mm}
\subsection{Effectiveness of Exploring Cognitive Features}
\label{exp:effectiveness_exploring_cognitive_features}
\begin{figure}[!htbp]
	\tiny
	\centering
	\includegraphics[width=0.5\textwidth]{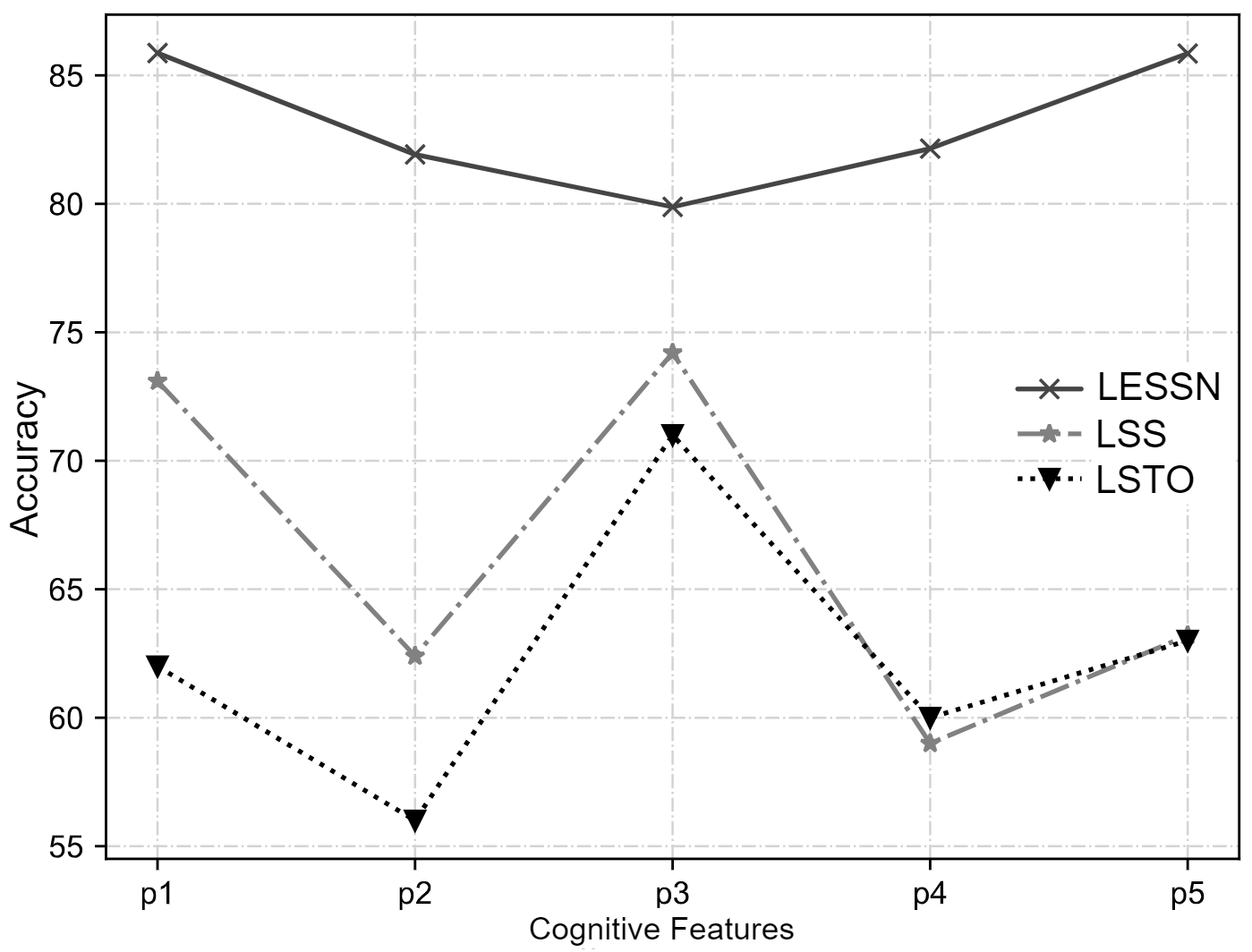}
	\vspace{-4mm}
	\caption{\small Performance of the exploring cognitive features models}
	\label{fig:exploring_cognitive_features_exp_plot}
	\vspace{-6mm}
\end{figure}

\vspace{-4mm}
As a prerequisite to the estimation of the professions, we need to examine how successfully one can exploit the cognitive cues from short-text vectors (section \ref{exp:effectiveness_exploring_cognitive_features}). Accordingly, we firstly compare the performance of the cognitive-semantic approaches. Our proposed model, named as LESSN, leverages five cognitive analysis modules including LIWC, Emoji, SPLICE, sentiStrength, and NRC (Section \ref{met:exploring_cognitive_features}). To investigate the performance of competitors, we use a cognitive dataset \citep{stillwell2015mypersonality} with 96-dimension feature space for both individual and combined attributes. We aim to discover the model that on the one hand, suits best the short-text noisy contents, and on the other hand, is the most competent candidate to infer cognitive features. We compare the LESSN model against LSS \citep{tadesse2018personality} and LSTO \citep{farnadi2013recognising}. The former is equipped with LIWC, SNA, and SPLICE, and the latter takes advantage of the LIWC, SNA, and other Time-related and auxiliary features. Figure \ref{fig:exploring_cognitive_features_exp_plot} compares the accuracy of the competitors where the dimension varies. The LESSN and LSTO respectively gain the highest and the lowest performance. The LESSN, as the most noise-resilient model, overcomes the LSS model because it can better extract the linguistic features in the training stage. The excessive noise in microblog contents can lead to hidden linguistic features that can significantly reduce the performance of LSS and LSTO. This is where the sentiStrength and NRC modules assure a better accuracy for the LESSN. The highest and the average accuracies for LSS in the estimation of the cognitive feature are respectively 74.85\% and 66.59\%, where the same numbers for LSTO are 71.38\% and 63.02\%. Our proposed framework achieves the highest accuracy of 86.10\% that is approx. 17\% more than \citep{tadesse2018personality}.
\vspace{-6mm}
\\
\subsection{Baselines}
\label{baselines}
\vspace{-4mm}
The baselines to infer professions are as follows:
\vspace{-2mm}
\begin{itemize}
	\item\textit{$Inprs_{cluster}$}: this method \citep{shafiabady2016using} uses SVM-Behavior to compute the weights and categorize job boundaries.
	\item\textit{$Inprs_{boost}$}: \citep{lu2019nonparametric} divides each cognitive feature by a single tree to obtain the boosting score.
	\item\textit{$Inprs_{curve}$}: is motivated by \citep{liu2018curve} and generates the curve weight via utilizing the ICF module.
	\item\textit{$Inprs_{conjunct}$}: is our proposed novel framework that adjusts parameters (Sec. \ref{met:coherence_weight}) to construct the $R^w$-tree that identifies cognitive profession boundaries.
	\item\textit{$Smart4Job$}: this time-aware approach \citep{benabderrahmane2017smart4job} combines the semantic classifier with the time series module to single-out the popularity of the occupations.
	\item\textit{$JobSeeker$}: this embedding model \citep{valverde2018job} automatically extracts the skills relating to each user profile.
\end{itemize}
\vspace{-4mm}
\begin{table}[h]
	\centering
	\caption{Baselines Compares}
	\vspace{-4mm}
	\label{tbl:baselines_compare}
	\def\arraystretch{1.5}
	\begin{tabular}{|l|l|l|l|}
		\hline
		\textbf{Methods} & \textbf{F-Measure} & \textbf{Precision} & \textbf{Recall}                                                                                                      
		\\ \hline
		\begin{tabular}[c]{@{}l@{}}{$Inprs_{cluster}$}\end{tabular} &
		\begin{tabular}[c]{@{}l@{}}0.711\end{tabular}   & \begin{tabular}[c]{@{}l@{}}0.770\end{tabular} &
		\begin{tabular}[c]{@{}l@{}}0.657\end{tabular}
		\\ \hline				
		\begin{tabular}[c]{@{}l@{}}{$Inprs_{boost}$}\end{tabular} &
		\begin{tabular}[c]{@{}l@{}}0.849\end{tabular}   & \begin{tabular}[c]{@{}l@{}}0.809\end{tabular} &
		\begin{tabular}[c]{@{}l@{}}0.890\end{tabular}
		\\ \hline
		\begin{tabular}[c]{@{}l@{}}{$Inprs_{curve}$}\end{tabular} &
		\begin{tabular}[c]{@{}l@{}}0.855\end{tabular}   & \begin{tabular}[c]{@{}l@{}}0.813\end{tabular} &
		\begin{tabular}[c]{@{}l@{}}0.898\end{tabular}
		\\ \hline
		\begin{tabular}[c]{@{}l@{}}{$Inprs_{conjunct}$}\end{tabular} &
		\begin{tabular}[c]{@{}l@{}}\textbf{0.890}\end{tabular}   & \begin{tabular}[c]{@{}l@{}}\textbf{0.858}\end{tabular} &
		\begin{tabular}[c]{@{}l@{}}\textbf{0.922}\end{tabular}
		\\ \hline
		\begin{tabular}[c]{@{}l@{}}{$Smart4Job$}\end{tabular} &
		\begin{tabular}[c]{@{}l@{}}0.865\end{tabular}   & \begin{tabular}[c]{@{}l@{}}0.849\end{tabular} &
		\begin{tabular}[c]{@{}l@{}}0.880\end{tabular}
		\\ \hline
		\begin{tabular}[c]{@{}l@{}}{$JobSeeker$}\end{tabular} &
		\begin{tabular}[c]{@{}l@{}}0.799\end{tabular}   & \begin{tabular}[c]{@{}l@{}}0.814\end{tabular} &
		\begin{tabular}[c]{@{}l@{}}0.783\end{tabular}
		\\ \hline
	\end{tabular}
\vspace{-4mm}
\end{table}
\vspace{-6mm}
\subsection{Benchmark}
\label{benchmark}
\vspace{-4mm}
Intuitively, we define statistical hypothesis parameters to evaluate the effectiveness of the competitors in inferring the professions. We investigate how the models succeed in assigning the right occupation to each of the microblog users. We dedicate 80\% of each dataset for parameter setting and perform the test on the remaining portion. For the ground-truth, we rely on two sources: The career tags in the Bio and the job labels that the experts add for the test users. The \textit{Recall} is the number of authors for whom we have successfully assigned a valid occupation. To compute the \textit{Precision} we divide the number of authors with correct jobs by the total number of users with the assigned profession tags. Finally, we judge the leading method using the \textit{F1-Score}, measured by Precision and Recall \citep{hosseini2019teags, khalatbari2019mcp}.
\vspace{-6mm}
\subsection{Effectiveness of the Profession Estimation}
\label{comparison_inferring_professions_methods}
\vspace{-4mm}
As Table \ref{tbl:baselines_compare} reports, this section compares the effectiveness of various competitors (Section \ref{baselines}) in inferring the professions using cognitive-semantics cues. For our tripartite framework, $Inprs_{curve}$ overcomes both $Inprs_{cluster}$ and $Inprs_{boost}$ because it tracks job-specific dynamic alterations through isotonic fitting. However, obtaining the best-adjusted parameters like the number of clusters remains a dilemma for $Inprs_{boost}$ and $Inprs_{cluster}$.  Employing the fixed margin to distinguish the career clusters, suddenly changed by an unobserved instance, makes $Inprs_{cluster}$ gain the least performance. The contribution of various aspects makes our framework fit to overcome other rivals.\\
Inherently, the quantity measure can alter in various pertinent boundaries. For instance, the domain knowledge of \textit{computer engineering} can conversely differ from the \textit{secretary}. Therefore, the more specific the profession, the higher the probability will be for the user to get assigned with a job, resulting in a higher recall. On the contrary, $Smart4Job$ can achieve better F-measure than $JobSeeker$ because it utilizes the semantic classification, directly increasing the domain knowledge, while $JobSeeker$ cannot compete merely via textual analysis tools. Hence, the variation between occupation boundaries is more authoritative than the textual representation of the job-specific tasks. Accordingly, the embedding approaches gain a lower performance compared to the isotonic regressors. Finally, due to the dataset-specific attributes, even the most straightforward boosting models surpasses the $Smart4Job$ that comprises both the classification and forecasting modules.
\vspace{-6mm}
 \subsection{Effectiveness of Short-text Contents History}
 \label{Effect_of_tweets_history_occupation_estimation}
\begin{figure}[h]
 	\tiny
 	\centering
 	\includegraphics[width=0.5\textwidth]{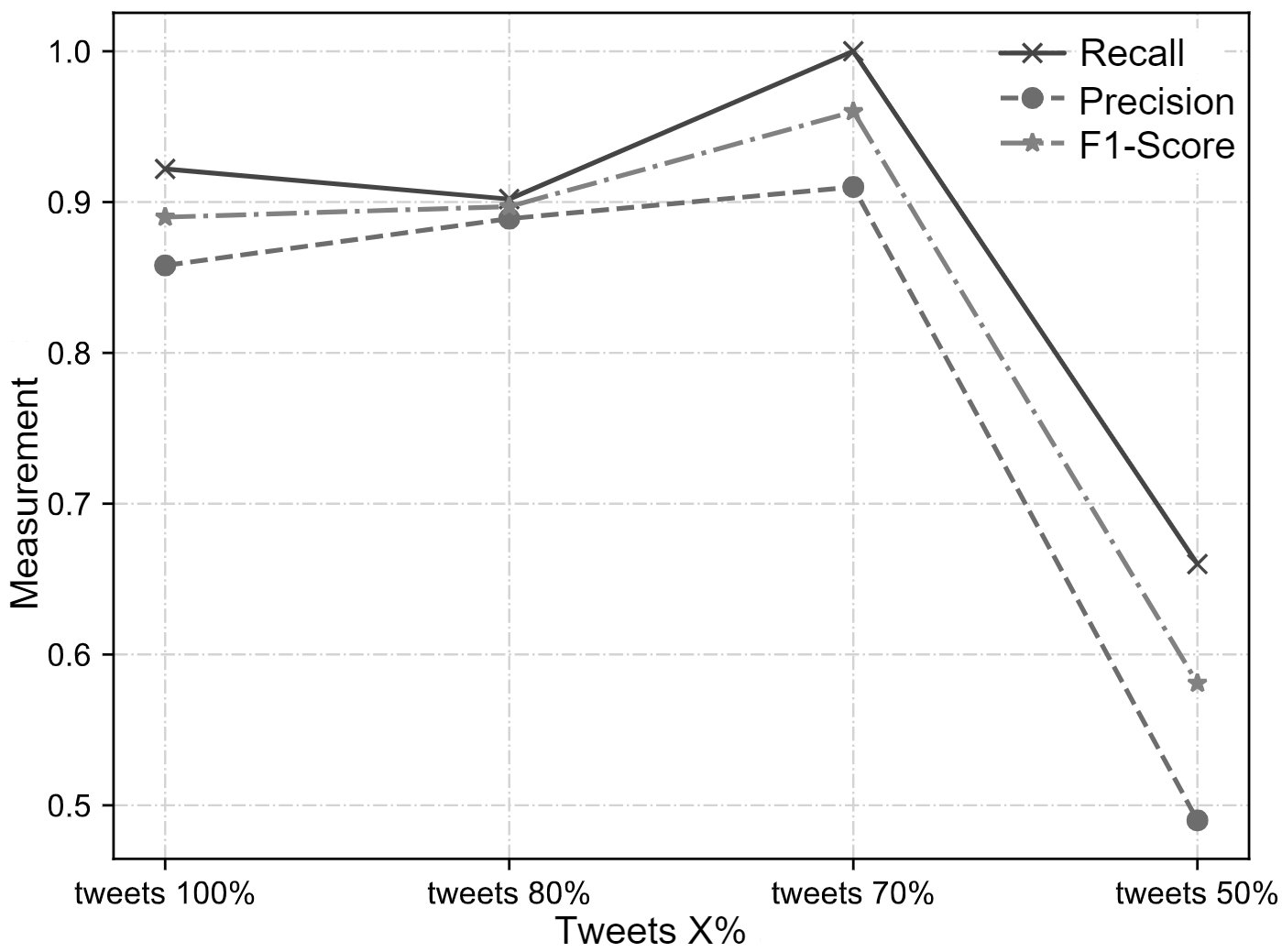}
 	\vspace{-3mm}
 	\caption{\small Impact of short-text history}
 	\label{fig:history_exp}
 	\vspace{-4mm}
 \end{figure}

 \vspace{-4mm}
 In this part, we examine how the data incompleteness can affect the performance of the career inference process. Firstly, we chronologically sort the tweet history of the users. We then eliminate a portion from older content and repeat the evaluation in each iteration.  Figure \ref{fig:history_exp} depicts the performance metrics through the removal of three ratios 20\%, 30\%, and 50\%.
 We observe that where we remove up to 30\% of the contents, disregarding the recall, which behaves overfitted, the precision increases continuously and we reach the best F1-Score in 70\%, i.e. 0.96. We may further improve the performance by removing the contents that are irrelevant to the cognitive cues, that is why the effectiveness of our proposed model fosters compared to the entire corpus. This can be due to the trends within data, where the older textual contents can mislead the pertaining cues of the jobs. Moreover, the noisy event-oriented sentences (e.g. about a new year) in the range of 100\%-80\% can negatively influence the recall metric. In a nutshell, more data does not assure a better result, especially when the amount of career-independent or event-oriented contents are abundant.
 \vspace{-5mm}
\vspace{-3mm}
\subsection{Impact of Parameter Adjustment}
\label{exp:impact_of_parameter_adjustment}
\begin{figure}[!htp]
	\tiny
	\centering
	\includegraphics[width=0.5\textwidth]{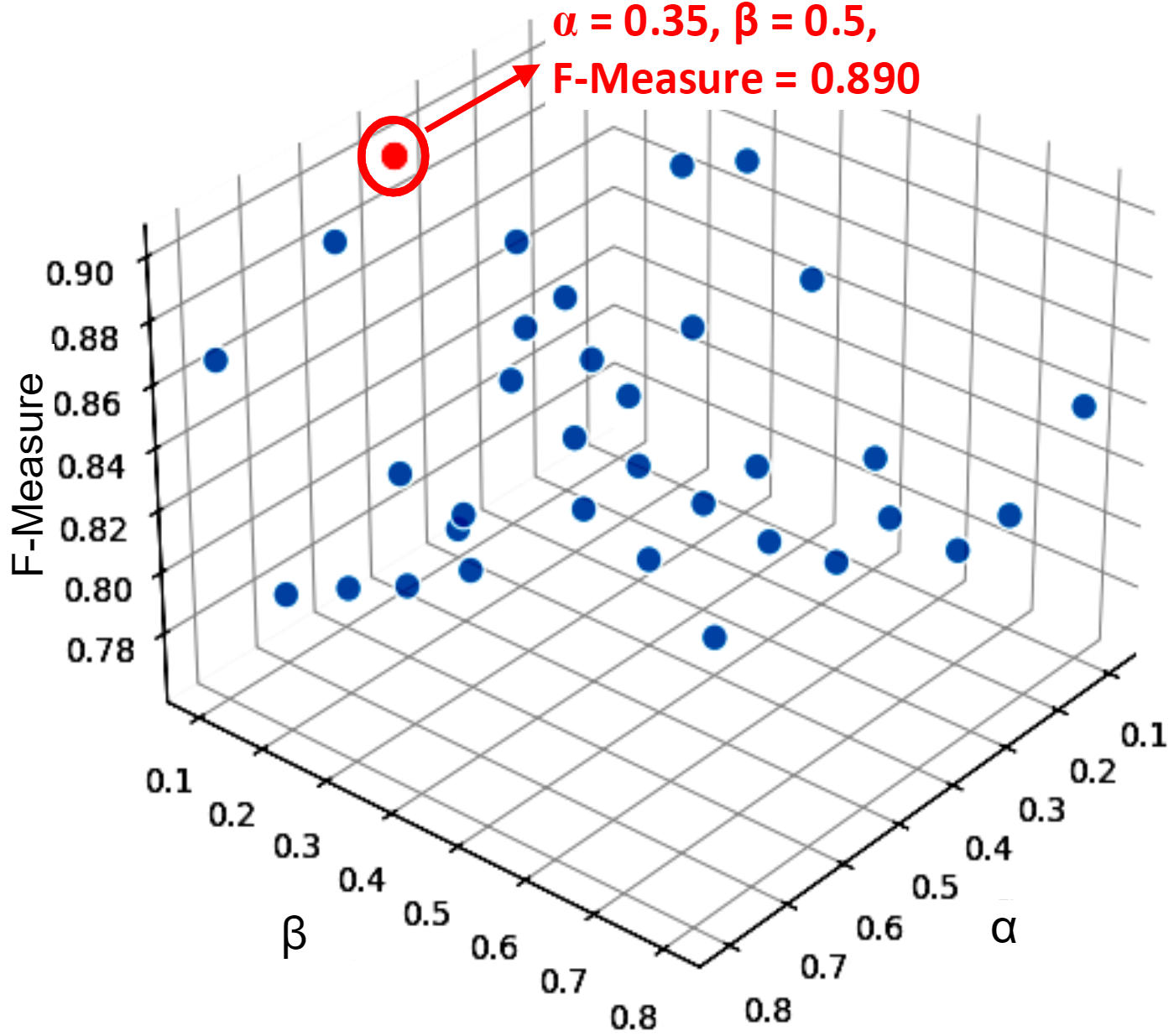}
	\vspace{-3mm}
	\caption{\small Impact of Parameters Adjustment}
	\label{fig:impact_of_adjustment_exp}	
	\vspace{-6mm}
\end{figure}

\vspace{-4mm}
It is inherently appealing to estimate professions by utilizing various data perspectives through different algorithms, where we incorporate three modules in our framework: (i) Cluster attribute ($\hat{w}_{cluster}$) as an SVM-Behavior module that considers the distribution impact of the cognitive feature clusters. (ii) Boost attribute($\hat{w}_{boost}$) that appoints a distinct decision tree for each cognitive cue. (iii) Curve attribute $\hat{w}_{curve}$ that avoids any negligence of the odd cognitive features through Isotonic Curve-Fitting. Nevertheless, the parameter adjustments can play a key role in promoting of our proposed framework $Inprs_{conjunct}$. Hence, we should include the impact of every module separately. In other words, the more optimized the parameters are, the better the performance will be.\\
\textbf{Adjusting $\alpha$ and $\beta$:} Figure \ref{fig:impact_of_adjustment_exp} illustrates a schematic of the tuning procedure to select the best values for $\alpha$ and $\beta$. We adopted performance metrics to ensure that our method could surpass other baselines. Concerning Eq. \ref{formula:coherence_weight}, we adjusted $\alpha$ and $\beta$, (ranging $\in$ [0,1]) to compute the best coherence score ($\hat{w}_C(P_i, o_j)$) for each profession $o_j$ based on the given input set of cognitive features $P_i$.
We used F1-Score to choose the best performance as it implicitly comprises both precision and recall measures. We further conducted a separate experiment for each pair of $\alpha$ and $\beta$ and used a random set of cognitive features. The best value for $\alpha$ and $\beta$ are respectively $0.35$ and $0.5$, which is reported for the best F1-score $0.890$.
\vspace{-7mm}
 \subsection{Impact of $R^w$-tree on Efficiency}
 \label{exp:efficiency_rtree}
\begin{figure}[htp]
 	\tiny
 	\centering
 	\includegraphics[width=0.5\textwidth]{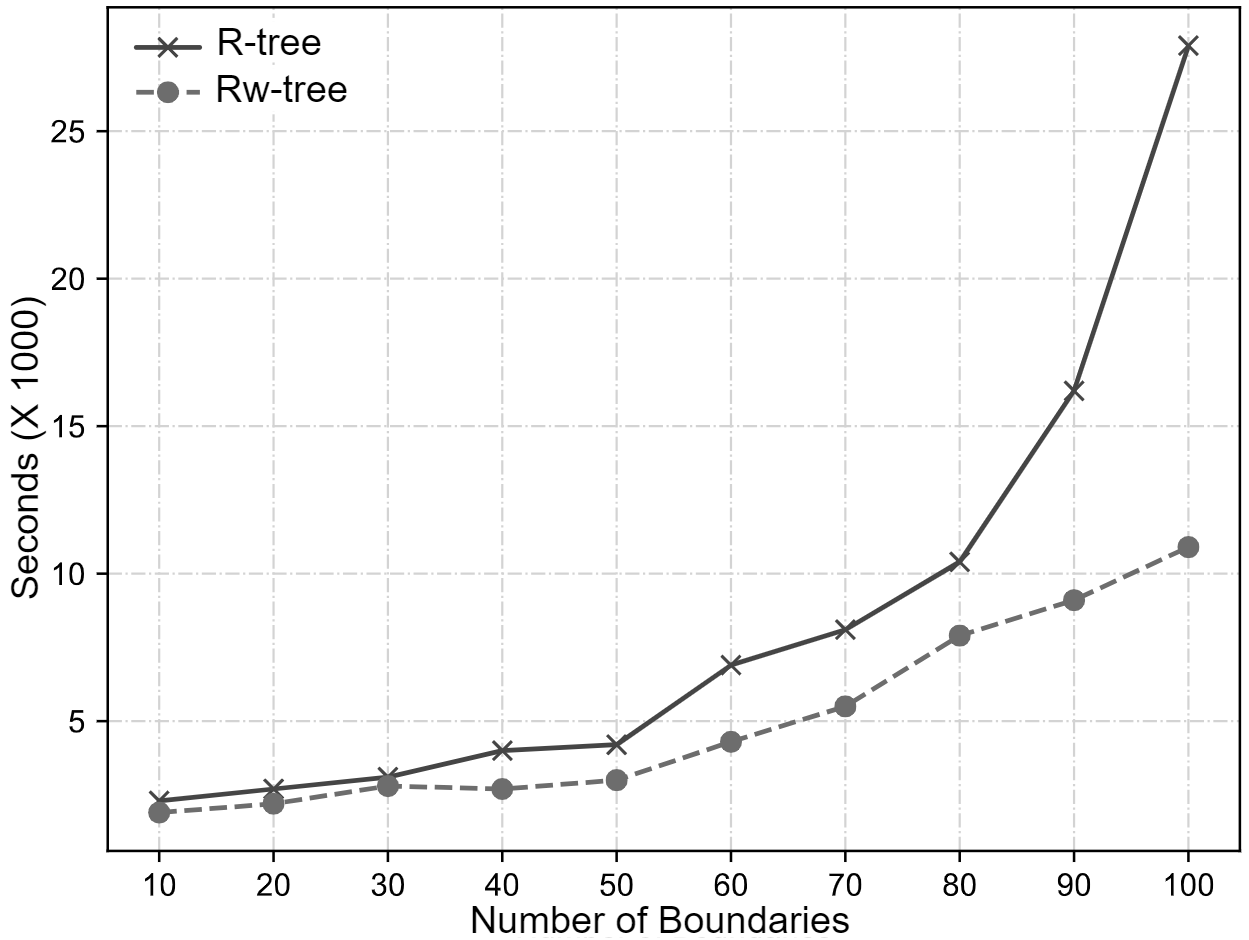}
 	\vspace{-3mm}
 	\caption{\small Compare Efficiency Between $R^w$-tree and $R$-tree}
 	\label{fig:exp_r-tree}
 	\vspace{-8mm}
 \end{figure}

 \vspace{-4mm}
 In this part, we compare the efficiency of $R^w$-tree versus original $R$-tree in construction of job boundaries. Fig. \ref{fig:exp_r-tree} depicts that as the value for boundaries increases, the latency of all two methods slightly grows. The reason is that choosing a bigger value for boundaries increases the size of the cognitive features which results in higher processing times.
 The results show that our model can efficiently exploit execution time equal to 11.2 ms for 100 boundaries, Where $R$-tree takes 27.5 millisecond. The reason is that $R^w$-tree separates the nodes with higher-weights from lower-weights in a separate layer and processes the operations for higher weight nodes and delay the rest.
 \vspace{-6mm}
\vspace{-9mm}
\section{Conclusion}
\label{sec:conclusion}
\vspace{-4mm}
In this work, we devise a unified cognitive-semantic framework that consumes short-text contents to predict career boundaries. To this end, we firstly extract linguistic features from temporal-textual data using the proposed textual analysis tool (LESSN), a better understanding of contextual semantics. More specifically, the LESSN model considers linguistic features to categorize the similarity between short-text vectors in semantic dimensions, for example, destruct and hate in the anger dimension, and then explores the cognitive features, such as openness, extraversion, and etcetera. Correspondingly, we attain the career boundaries using tripartite weighting modules such as clustering, boosting, and isotonic fitting that collectively learn how every cognitive orient is prevalent to the given careers. Consequently, where the coherence weight maximizes the performance, we construct a novel $R^w$-tree index hierarchy to adapt each cognitive-semantic category to the given agglomerative profession boundaries.
The extensive experiments on a real-world microblog dataset demonstrate the superiority of our proposed model versus trending competitors. While we can leverage the deep learning algorithms to improve the effectiveness, we can also scale-out the components to promote efficiency, parallel processing. We leave these tasks as future work.
\vspace{-8mm}
\bibliographystyle{spbasic}      
\bibliography{IEEEexample}   

\end{document}